\documentclass[10pt,reqno]{amsart}

\usepackage{companypaper}
\usepackage{algorithm}
\usepackage{algorithmic}
\usepackage{natbib}
\usepackage{hyphenat}
\usepackage{enumitem}
\usepackage{verbatim}

\usepackage{amsmath}
\usepackage{multirow}
\usepackage{makecell}
\usepackage{caption}
\usepackage{amssymb}

\usepackage{tabularx}
\usepackage{array}
\usepackage{booktabs}
\newcommand{\norm}[1]{\left\lVert #1 \right\rVert_2}
\usepackage[most]{tcolorbox}

\newtcolorbox{promptbox}{
  colback=gray!5,
  colframe=black!75,
  boxrule=0.5pt,
  arc=2pt,
  left=6pt,
  right=6pt,
  top=6pt,
  bottom=6pt,
  breakable=false
}
\makeatletter
\pretocmd{\@settitle}{\let\uppercasenonmath\@gobble}{}{}
\patchcmd{\@settitle}{\bfseries}{\bfseries\LARGE}{}{}
\pretocmd{\@setauthors}{\let\MakeUppercase\@firstofone}{}{}
\patchcmd{\@setauthors}{\centering\footnotesize}{\centering\large}{}{}
\apptocmd{\@setauthors}{}{}{}
\renewcommand{\@setaddresses}{}
\renewcommand{\@setdate}{%
  \noindent\normalfont\footnotesize
  \textsuperscript{1}Fujitsu Limited\par
  \vskip 2pt
  \textbf{Correspondence}: Ganesh Senrayan at \textcolor{blue}{\texttt{ganesh.senrayan@fujitsu.com}}\par
}
\makeatother

\CompanyLogo{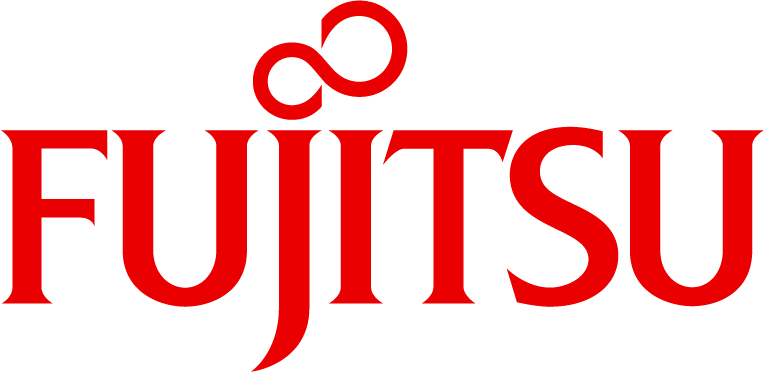}
\CompanyLogoHeight{11mm}
\CompanyHeaderText{Fujitsu Limited}
\CompanyDocType{}
\CompanyClassification{}
\CompanySubtitle{}
\CompanyRunningTitle{Exploratory and Assimilating Reflection for Long-term Memory}

\title{Exploratory and Assimilating Reflection: \\Reflective Recall Cycle for Long-term Memory}

\author{
Ganesh Senrayan\textsuperscript{1},
Moyuru Yamada\textsuperscript{1},
Ishan Jindal\textsuperscript{1},
Kiran Purohit\textsuperscript{1}}
\date{July 20, 2026}

\begin{document}

\maketitle

\begin{abstract}
LLM-based autonomous agents require external memory to overcome their statelessness and limited context window for long-term interaction and dynamic knowledge reasoning. However, existing memory retrieval methods often lack adaptability and sample efficiency, and struggle to retrieve the right mixture of memories from heterogeneous stores. We propose \textit{Exploratory-Assimilating Reflection (EAR)}, a framework for high initial retrieval performance and sample-efficient adaptation. EAR combines two mechanisms: Exploratory Reflection, which performs iterative search to bootstrap retrieval and collect useful experiences for each query, and Assimilating Reflection, which replays these experiences from an Experience Buffer to refine a global reranker more efficiently than methods relying only on immediate rewards. Experiments show that EAR improves retrieval by up to 17.9\% over the baseline retriever on two long-term dialogue benchmarks. We also show that EAR is highly sample-efficient and robust to noisy feedback.

\end{abstract}

\section{Introduction}
Autonomous agents powered by Large Language Models (LLMs) hold immense potential for automating complex, real-world tasks \citep{augmented_lm_survey}. However, the underlying architecture of LLMs is inherently stateless \citep{rmm}; their inability to consider context beyond a finite window makes it challenging to maintain long-term memory for personalized interaction or to reason over dynamically evolving knowledge bases. Thus, an external memory mechanism is crucial for the agents to overcome this fundamental limitation.

\begin{figure}[t]
    \centering
    \includegraphics[width=0.6\linewidth]{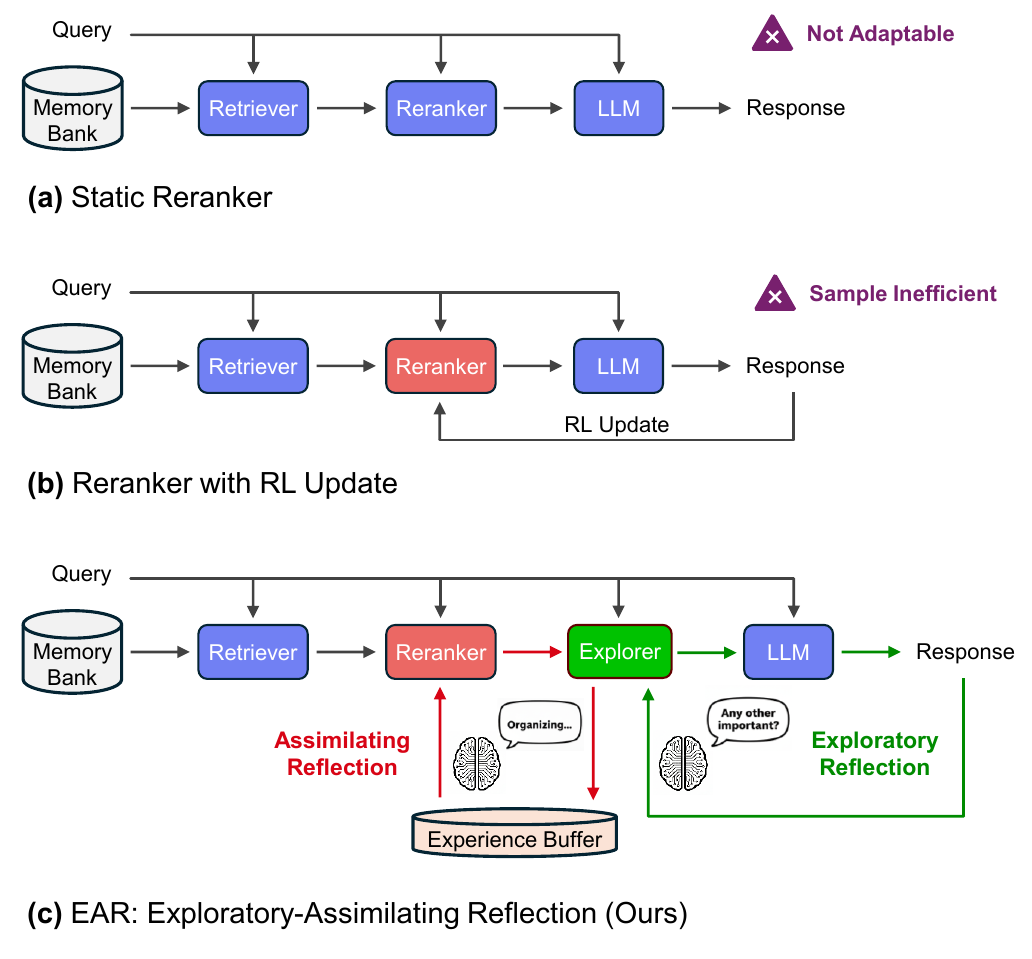}
    \caption{
    Comparison of memory retrieval paradigms. (a) Static reranker is unable to adapt to data distribution shifts. (b) Reranker with RL update struggles with sample inefficiency. (c) Our EAR (Exploratory-Assimilating Reflection) achieves both high initial recall and sample-efficient updates.
    }
    \label{fig:comparison}
\end{figure}

Surfacing the essential information for a query from a vast memory bank is a deceptively challenging task.
The most standard approach employs a static retriever, pre-trained on large-scale data, to identify relevant memory chunks based on semantic similarity in an embedding space \citep{rag}. It relies heavily on the pre-trained retriever, which leads to significant performance degradation when applied to out-of-domain tasks \citep{thakur2021beirheterogenousbenchmarkzeroshot}.
A common approach to mitigate this limitation is to introduce a reranker that is tuned offline \citep{rerank}. However, this offline-tuning approach, has two major drawbacks. It demands expensive data collection and fails to adapt to shifts in data distribution over time \citep{grundel2024galapagos, alkhalifa2024longeval}. (Fig. \ref{fig:comparison} (a)).


To adapt to domain shift, \citep{rmm} have introduced online adaptation method, which employs Reinforcement Learning (RL) to learn from LLM feedback during interactions and adapt to evolving domains and user patterns (Fig. \ref{fig:comparison} (b)). 
While this approach marks a significant step towards more adaptive and robust agents, the slow adaptation and instability inherent in its RL update makes it sample inefficient. This requires the agent to undergo a substantial number of interactions to improve, leaving it vulnerable to poor performance during the crucial initial stages.


To address the aforementioned limitations, we propose \textbf{Exploratory-Assimilating Reflection (EAR)}, a novel framework for adaptive memory retrieval.
For a given query, Exploratory Reflection performs an active, iterative search for the most relevant memory entries using an Explorer module. This search not only constructs a rich set of evidences for the given query but also produces valuable experiences for long-term adaptation.
Assimilating Reflection is designed to achieve the long-term adaptation of the global adapter module. Unlike \citep{rmm} that learns myopically from immediate rewards, Assimilating Reflection utilizes an Experience Buffer to store a history of exploratory experiences. By replaying and learning from this diverse set of past experiences, we efficiently and continuously update the reranker model, improving its overall performance over time.

Our framework is inspired by cognitive development. In humans, short-term memory (or working memory) maintains a limited amount of information for immediate use and long-term memory consolidates experiences over extended timescales \citep{baddeley2003working, squire2004memory}.  Cognitive theories suggest that rehearsal or replay enables transfer from short-term to long-term storage, and that long-term traces can be reactivated when needed \citep{danker2010ghosts}. Inspired by this interplay, EAR unifies query-specific evidence construction with continual model adaptation.


Extensive experiments on the LongMemEval and LoCoMo benchmarks demonstrate that our approach achieves significant improvements in recall, with absolute gains of up to 17.9\% and 12.5\% over the baseline retriever. Furthermore, EAR exhibits exceptional sample efficiency and remains highly robust to noisy feedback, validating the efficacy of its core components.
To test EAR's capability to adapt to a heterogeneous memory environment, we setup a case study on the LoCoMo dataset where a mixture of dialogue (\textit{episodic}) and observational (\textit{semantic}) evidence is available. We demonstrate that EAR consistently selects a more effective, balance of memory types. This balancing improves downstream answering accuracy by to 13.6\%, proving that EAR’s adaptivity seamlessly extends to mixing distinct memory stores.


Our contributions are summarized as follows:
\begin{itemize}
    \item We propose Exploratory-Assimilating Reflection (EAR), a novel framework for adaptive memory retrieval that combines iterative selection (Explorer) with long-term adaptation from past interactions (Experience Buffer).
    \item We highlight EAR's superior sample efficiency, demonstrating that it reaches static baseline performance in 77\% fewer feedback steps than existing RL-based retrieval method.
    \item We show that EAR seamlessly balances heterogeneous memory sources. When presented with multiple memory stores, it adaptively retrieves an optimal, mixture of evidence.
\end{itemize}
\begin{figure*}[ht]
    \centering
    \includegraphics[width=1.0\linewidth]{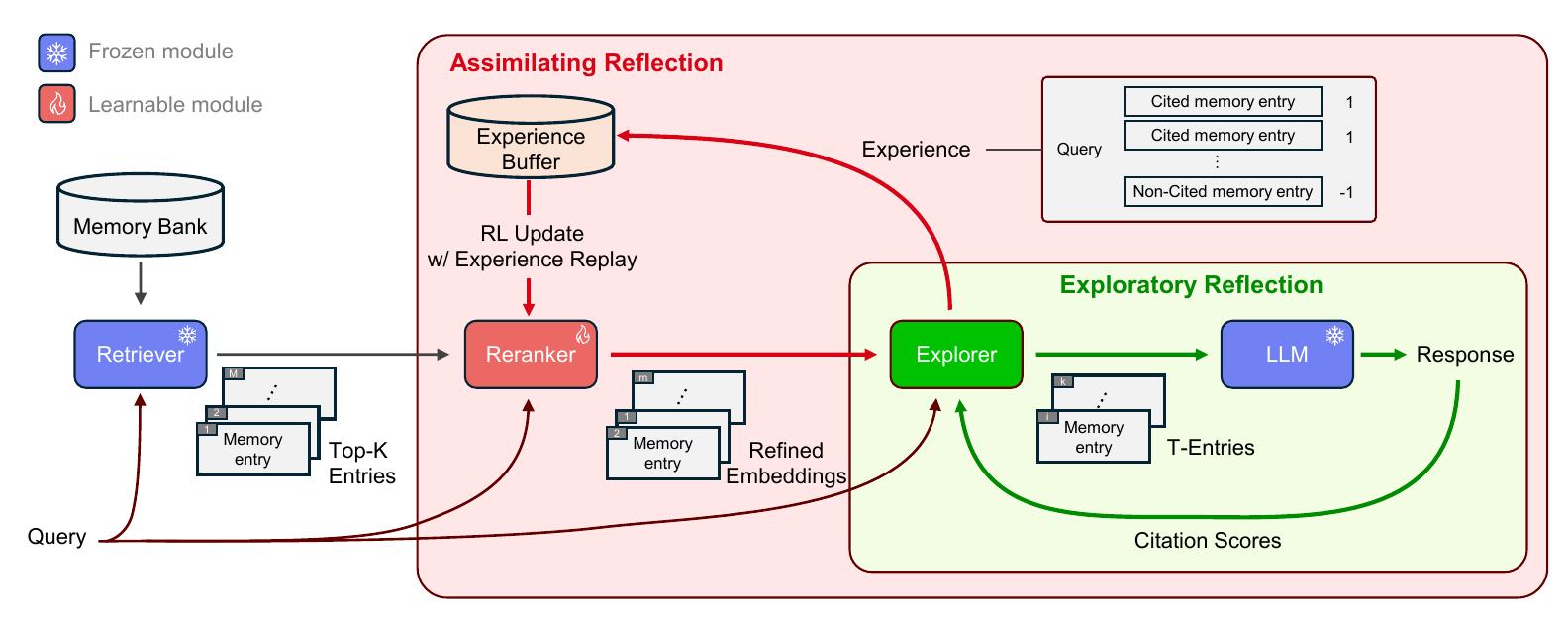}
    \caption{Overview of Exploratory-Assimilating Reflection (EAR) Framework. Exploratory Reflection performs an iterative deep search within a single query to build a rich set of useful context using Explorer, thereby bootstrapping the learning process of the adaptive reranker. Assimilating Reflection takes a broad view across a history of these explorations stored in Experience Buffer, efficiently improve the global reranker model.}
    \label{fig:pipeline}
\end{figure*}

\section{Related Work}
Our framework aims to equip LLM-based agents with a robust, continually adapting memory system. To contextualize our contributions, we study three interconnected areas of prior work: (1) LLM-based autonomous agents, (2) standard retrieval-augmented reasoning pipelines used to overcome their inherent statelessness, and (3) recent adaptive memory retrieval methods, where we identify the critical sample-efficiency gaps addressed by EAR.
\paragraph{LLM-based Autonomous Agents}
The advent of powerful Large Language Models (LLMs) such as GPT-4 \citep{gpt4} and Llama 3 \citep{Llama3} has led to the development of sophisticated autonomous agents. These agents aim to perform complex, multi-step tasks by leveraging the reasoning and planning capabilities of LLMs. Seminal works have explored various agent architectures, including those that use tools \citep{toolformer}, generate task-specific plans \citep{react}, or engage in self-reflection to improve their performance \citep{reflexion}.
However, a fundamental limitation of these agents is their reliance on the LLM's finite context window and static parametric knowledge. 
To this end, several architectures have been proposed to explicitly manage long-term memory \citep{memorybank,ld-agent}, which are designed to maintain conversational consistency over extended interactions. 
Thus, the mechanism by which an agent retrieves information from its memory is a critical component that directly governs efficacy and intelligence.
\paragraph{Retrieval-Augmented Reasoning}
Retrieval-Augmented Generation (RAG) \citep{rag}, has become a standard for supplementing LLMs with external knowledge. 
To improve the retrieval performance, subsequent works use a efficient retrievers (sparse retriever like BM25 \citep{bm25}, dense retrievers like Contriever \citep{contriever}) to fetch candidate documents followed by a powerful reranker \citep{rerank}.

While a finetuned reranker is effective for static knowledge bases, it faces significant challenges when applied to the long-term memory of an autonomous agent.
Furthermore, a memory retrieval mechanism for autonomous agents must be robust and adaptable to the diverse retrieval demands of varying conversation domains and individual user interaction patterns.
Methods trained on a fixed corpus struggle to handle this dynamic nature, necessitating adaptive methods.
\paragraph{Adaptive Memory Retrieval}
More recent and sophisticated approaches aim to make the retrieval process itself more dynamic and intelligent at inference time. For instance, Self-RAG \citep{self-rag} fine-tunes an LLM to generate special "reflection tokens" that control the retrieval process. 
Similarly, approaches like Adaptive-RAG \citep{adaptive-rag} learn a policy to adapt the retrieval strategy based on the query's complexity.
While these methods make the retrieval decision process highly adaptive, the control policies themselves are typically trained offline on a fixed dataset. They are not designed to continually update based on the feedback from ongoing agent interactions.
The need for robust and adaptive retrieval is particularly important in agents designed for long-term interaction, such as MemoryBank \citep{memorybank} and LD-Agent \citep{ld-agent}. 
To address the need for continual learning, the most recent method has proposed a method to update the reranker based on feedback signals from the LLM through REINFORCE algorithm \citep{rmm}. This method marks a significant step towards more adaptive agents. However, it suffers from critical sample inefficiency, requiring a substantial number of interactions to learn effectively. 
\section{Methodology}
Our proposed framework shown in Figure \ref{fig:pipeline} is composed of two distinct modules:
\textbf{Exploratory Reflection} acts as a short-term memory, performing responsive, query-specific refinement. 
\textbf{Assimilating Reflection} serves as a long-term memory, integrating feedback across queries to adapt the global embedding space. 
Synergistically, they form a reflection cycle that balances immediate adaptivity and cumulative learning, rendering the process more sample-efficient and cognitively plausible. 

\subsection{Problem Formulation}

We consider an online adaptive memory retrieval setting. At each
interaction, the system receives a query, retrieves a finite candidate set from an external memory bank, and selects a small subset of memories for downstream answer generation. The selected memories are evaluated by an external critic, and the resulting feedback is used to continually adapt the retrieval model.

For clarity, we first formulate EAR for a single generic query and suppress the online interaction index. We introduce the interaction index \(n\) only when describing continual adaptation and experience replay.

\label{sec:problem-formulation}

Let \(x\) denote a textual query and let
    $\mathcal{V} =
    \left\{
        v_1,\ldots,v_M
    \right\}$
denote an external memory bank containing \(M\) textual memory entries.
A frozen encoder \(f_{\theta}\) maps the query and memories to
\(d\)-dimensional embeddings:
$
    \mathbf{q}
    =
    f_{\theta}(x)
    \in
    \mathbb{R}^{d},
    \quad
    \mathbf{m}_i
    =
    f_{\theta}(v_i)
    \in
    \mathbb{R}^{d}.
    \label{eq:base-embeddings}
$
The frozen first-stage retriever selects the \(K\) memories with the
highest query--memory similarity:
\begin{equation}
    \mathcal{C}
    =
    \operatorname*{TopK}_{i \in [M]}
    \operatorname{sim}
    \left(
        \mathbf{q},
        \mathbf{m}_i
    \right),
    \qquad
    |\mathcal{C}| = K,
    \label{eq:candidate-retrieval}
\end{equation}
where \([M]=\{1,\ldots,M\}\) and
\(\operatorname{sim}(\cdot,\cdot)\) denotes cosine similarity. From the candidate set \(\mathcal{C}\), the retrieval module constructs a slate of memories $\mathcal{S}$ from $\mathcal{C}$.
After the slate is evaluated by an external critic, feedback is observed
for each selected memory $    y_i \in \{-1,+1\},
    \quad
    i \in \mathcal{S}, $
where \(+1\) indicates that memory \(v_i\) is useful for answering the
query, and \(-1\) indicates that it is not useful. More generally, \(y_{i}\) can be interpreted as a noisy observation
of an unknown query--memory utility function
\(u(x,v_{i})\). The learning objective is to maximize the expected utility of selected
memories:
\begin{equation}
    J(\phi)
    =
    \mathbb{E}_{x\sim\mathcal{D}}
    \left[
        \frac{1}{|\mathcal{S}_\phi|}
        \sum_{i \in \mathcal{S}_\phi}
        u(x,v_i)
    \right],
    \label{eq:expected-utility}
\end{equation}
where \(\mathcal{D}\) denotes the distribution of incoming queries, \(u(x,v_i)\) is the unknown query--memory utility and
\(\phi\) denotes trainable adapter parameters.

\subsection{EAR Framework}

The Exploratory-Assimilating Reflection (EAR) framework consists of two synergistic components namely, \textbf{Exploratory Reflection} and \textbf{Assimilating Reflection}. The overall two-step process is summarized in Algorithm \ref{alg:ear_main}, with detailed component-level algorithms deferred to \textbf{Appendix \ref{app:implementation}}.

\begin{algorithm}[tb]
\small
\caption{Exploratory-Assimilating Reflection (EAR)}
\label{alg:ear_main}

\begin{algorithmic}[1]
\item[] \raggedright \textbf{Input}: Query $x$, memory bank $\mathcal{V}$
\item[] \raggedright \textbf{Output}: Memory subset $\mathcal{S}_T$
\STATE $q, \mathcal{C} \leftarrow \text{f}_{\theta}(x, \mathcal{V})$ \hfill \textit{// Retrieve top-k}
\STATE $q', \mathcal{C}' \leftarrow \text{g}_{\phi}(q, \mathcal{C})$ \hfill \textit{// Adapt embeddings}
\FOR{$t = 1$ to $T$}
    \STATE $\mathcal{S}_t \leftarrow \text{BanditSelect}(q', \mathcal{C}')$ \hfill \textit{//Exploratory Reflection} 
    \STATE $\mathbf{y}_t \leftarrow \text{CriticFeedback}(\mathcal{S}_t)$ \hfill \textit{// $\mathbf{y}_t \in \{-1, +1\}^{|\mathcal{S}_t|}$}
    \STATE $\text{UpdateUCBStats}(\mathbf{y}_t)$ 
\ENDFOR
\STATE $\bar{\mathbf{y}} \leftarrow \frac{1}{T} \sum_{t=1}^T \mathbf{y}_t$ \hfill \textit{// Aggregate rewards}
\STATE $\mathcal{B} \leftarrow \mathcal{B} \cup \{(q, \mathcal{C}, \bar{\mathbf{y}})\}$ \hfill \textit{// Experience Storage}
\STATE $b \sim \mathcal{B}$ \hfill \textit{// Sample replay mini-batch}\\

\STATE $\phi \leftarrow \phi - \eta \nabla_\phi \big( \mathcal{L}_{curr}(\phi) + \mathcal{L}_{replay}(\phi; b) \big)$  \hfill \textit{// Assimilating Reflection}
\STATE \textbf{return} \ensuremath{\mathcal{S}_T} \strut
\end{algorithmic}
\end{algorithm}



\subsection{Residual Retrieval Adapter}
\label{sec:adapter}

EAR learns residual transformations for the query and memory
embeddings. Let
$
    \phi
    =
    \left\{
        \mathbf{W}_{q},
        \mathbf{W}_{m}
    \right\},
    \mathbf{W}_{q},
    \mathbf{W}_{m}
    \in
    \mathbb{R}^{d \times d}.$

The adapted query and memory embeddings are
\begin{equation}
    \widetilde{\mathbf{q}}
    =
    \frac{
        \left(
            \mathbf{I}+\mathbf{W}_{q}
        \right)
        \mathbf{q}
    }{
        \norm{
            \left(
                \mathbf{I}+\mathbf{W}_{q}
            \right)
            \mathbf{q}}
    }
    \quad
     \widetilde{\mathbf{m}}_i
    =
    \frac{
        \left(
            \mathbf{I}+\mathbf{W}_{m}
        \right)
        \mathbf{m}_i
    }{
        \norm{
            \left(
                \mathbf{I}+\mathbf{W}_{m}
            \right)
            \mathbf{m}_i
        }
    }
    \label{eq:query-adapter}
\end{equation}

The adapter defines the following distribution over the candidate memories $\mathcal{C}$:
\begin{equation}
    p_{\phi}
    \left(
        i
        \mid
        x,\mathcal{C}
    \right)
    =
    \frac{
        \exp\left(z_i^{\phi}\right)
    }{
        \displaystyle
        \sum_{\ell\in\mathcal{C}}
        \exp\left(z_{\ell}^{\phi}\right)
    };
    \quad
        z_i^{\phi} = \widetilde{\mathbf{q}}^{\top} \widetilde{\mathbf{m}}_i, 
    \quad
    i\in\mathcal{C}.
    \label{eq:adapter-distribution}
\end{equation}


\subsubsection{Exploratory Reflection}
\label{subsec:exploratory_reflection}

Exploratory Reflection performs \(T\) query-specific exploration rounds, each constructing a slate of \(s\) memories from the candidate set \(\mathcal{C}\). Each candidate memory \(i\in\mathcal{C}\) is treated as
a query-local bandit arm. Its selection count \(N_i^{(t)}\) and empirical
mean feedback \(\widehat{\mu}_i^{(t)}\) are initialized as
$    N_i^{(0)} = 0,
    \widehat{\mu}_i^{(0)} = 0.
    \label{eq:bandit-initialization}
$
At exploration round \(t\), the upper-confidence value of candidate
\(i\) is
\begin{equation}
    U_i^{(t)}
    =
    \widehat{\mu}_i^{(t-1)}
    +
    \alpha
    \sqrt{
        \frac{
            \log\!\left(
                1+\sum_{\ell\in\mathcal{C}}N_\ell^{(t-1)}
            \right)
        }{
            1+N_i^{(t-1)}
        }
    },
    \label{eq:ucb}
\end{equation}
where \(\alpha\geq0\) controls the exploration strength.

Let
\(\mathcal{S}_{j-1}^{(t)}
=\{i_1^{(t)},\ldots,i_{j-1}^{(t)}\}\)
denote the partial slate before selecting its \(j\)-th memory. The
redundancy of candidate \(i\) with the already selected memories is
\begin{equation}
    D_i^{(t,j)}
    =
    \begin{cases}
        0,
        & j=1,
        \\[3pt]
        \displaystyle
        \max_{\ell\in\mathcal{S}_{j-1}^{(t)}}
        \widetilde{\mathbf{m}}_i^\top
        \widetilde{\mathbf{m}}_\ell,
        & j>1.
    \end{cases}
    \label{eq:diversity-penalty}
\end{equation}

The Explorer combines adapted query--memory relevance, slate diversity,
and uncertainty:
\begin{equation}
    A_i^{(t,j)}
    =
    \lambda_{\mathrm{rel}}
    \widetilde{\mathbf{q}}^\top\widetilde{\mathbf{m}}_i
    -
    \lambda_{\mathrm{div}}D_i^{(t,j)}
    +
    \lambda_{\mathrm{exp}}U_i^{(t)}.
    \label{eq:explorer-score}
\end{equation}

The \(j\)-th memory is selected without replacement as
\begin{equation}
    i_j^{(t)}
    =
    \operatorname*{arg\,max}_{
        i\in\mathcal{C}\setminus\mathcal{S}_{j-1}^{(t)}
    }
    A_i^{(t,j)}.
    \label{eq:explorer-selection}
\end{equation}
Repeating Equation~\eqref{eq:explorer-selection} for
\(j=1,\ldots,s\) produces the exploration subset
$
    \mathcal{S}^{(t)}
    =
    \left\{
        i_1^{(t)},\ldots,i_s^{(t)}
    \right\}.
$

The external critic then assigns each selected memory a usefulness label
$
    y_i^{(t)}\in\{-1,+1\},
    i\in\mathcal{S}^{(t)}.
    \label{eq:exploration-feedback}
$
For every selected candidate \(i\in\mathcal{S}^{(t)}\), its bandit
statistics are updated as
\begin{equation}
    N_i^{(t)}
    =
    N_i^{(t-1)}+1,
    \quad
    \widehat{\mu}_i^{(t)}
    =
    \widehat{\mu}_i^{(t-1)}
    +
    \frac{
        y_i^{(t)}-\widehat{\mu}_i^{(t-1)}
    }{
        N_i^{(t)}
    }.
    \label{eq:bandit-update}
\end{equation}
The statistics of unselected candidates remain unchanged. After \(T\)
exploration rounds, EAR returns the final slate \(\mathcal{S} ^{\mathrm{EAR}} = \mathcal{S}^{(T)}\).

\subsection{Learning from Current-Query Feedback}
\label{sec:current-query-learning}

The Explorer in Equation~\eqref{eq:explorer-selection} performs
deterministic selection using relevance, diversity, and uncertainty.
We therefore optimize the adapter using a feedback-weighted ranking
objective.

For each selected memory, the adapter probability is computed
sequentially using the memories selected at previous slate positions.
The current-query loss is
\begin{equation}
    \mathcal{L}_{\mathrm{cur}}(\phi)
    =
    -
    \frac{1}{Ts}
    \sum_{t=1}^{T}
    \sum_{j=1}^{s}
    \left(
        y_{i_j^{(t)}}^{(t)}
        -
        b
    \right)
    \times
    \log
    p_{\phi}
    \left(
        i_j^{(t)}
        \mid
        x,
        \mathcal{C}
    \right),
    \label{eq:current-query-loss}
\end{equation}
where \(b\) is an action-independent baseline.


Minimizing Equation~\eqref{eq:current-query-loss} increases the
probability assigned to positively rewarded memories and decreases the
probability assigned to negatively rewarded memories.
\subsection{Assimilating Reflection with Experience Replay}
\label{sec:assimilating-reflection}

Exploratory Reflection provides query-specific feedback, whereas
Assimilating Reflection consolidates this feedback into persistent
updates of the global adapter. We maintain an Experience Buffer
containing feedback collected from previously processed queries.

For the \(n\)-th query, let
\(o_i^{(n,t)}=\mathbb{I}[i\in\mathcal{S}^{(n,t)}]\) indicate whether
candidate \(i\) was selected during exploration round \(t\). The set of
candidates for which feedback was observed is
\begin{equation}
    \mathcal{O}^{(n)}
    =
    \left\{
        i\in\mathcal{C}^{(n)}
        :
        \sum_{t=1}^{T}o_i^{(n,t)}>0
    \right\}.
    \label{eq:observed-memory-set}
\end{equation}

For each \(i\in\mathcal{O}^{(n)}\), we aggregate its feedback across
exploration rounds:
\begin{equation}
    \overline{y}_i^{(n)}
    =
    \frac{
        \sum_{t=1}^{T}
        o_i^{(n,t)}y_i^{(n,t)}
    }{
        \sum_{t=1}^{T}
        o_i^{(n,t)}
    }.
    \label{eq:aggregated-feedback}
\end{equation}
For replay, the aggregated feedback is converted to a binary reward \(
    \widetilde{y}_i
    =
    \operatorname{sign}(\overline{y}_i),
\) with ties mapped to \(-1\).
The resulting experience is stored as
\begin{equation}
    e^{(n)}
    =
    \left(
        \mathbf{q}^{(n)},
        \left\{
            \mathbf{m}_i^{(n)}
        \right\}_{i\in\mathcal{C}^{(n)}},
        \mathcal{O}^{(n)},
        \left\{
            \widetilde{y}_i^{(n)}
        \right\}_{i\in\mathcal{O}^{(n)}}
    \right),
\end{equation}

\begin{equation}
    \mathcal{B}_n
    =
    \mathcal{B}_{n-1}\cup\{e^{(n)}\}.
\end{equation}

To replay relevant past experiences, we select the \(B\) buffer entries
whose pre-adapter query embeddings are most similar to current
query:
\begin{equation}
    \mathcal{J}_n
    =
    \operatorname*{TopB}_{r<n}
    \operatorname{sim}
    \left(
        \mathbf{q}^{(n)},
        \mathbf{q}^{(r)}
    \right).
    \label{eq:replay-selection}
\end{equation}
For each replayed experience \(r\in\mathcal{J}_n\), we reapply the
current adapter and optimize the same feedback-weighted ranking
objective used for the current query:
\begin{equation}
    \mathcal{L}^{(n)}_{\mathrm{rep}}(\phi)
    =
    \frac{1}{|\mathcal{J}_n|}
    \sum_{r\in\mathcal{J}_n}
    \mathcal{L}_{\mathrm{rank}}
    \left(
        \phi;
        e^{(r)}
    \right),
    \label{eq:replay-loss-compact}
\end{equation}
where \(\mathcal{L}_{\mathrm{rank}}\) is defined in
Equation~\eqref{eq:current-query-loss}. The complete replay and
stochastic slate-sampling procedure is provided in
Appendix~\ref{app:implementation}.

Finally, the adapter is trained using both the current-query feedback
and replayed experiences:
\begin{equation}
    \mathcal{L}^{(n)}_{\mathrm{EAR}}(\phi)
    =
    \mathcal{L}^{(n)}_{\mathrm{cur}}
    \left(
        \phi
    \right)
    +
    \lambda_{\mathrm{rep}}
    \mathcal{L}^{(n)}_{\mathrm{rep}}(\phi),
    \label{eq:ear-loss}
\end{equation}
followed by the online update
\begin{equation}
    \phi_{n+1}
    =
    \phi_n
    -
    \eta
    \left.
        \nabla_{\phi}
        \mathcal{L}^{(n)}_{\mathrm{EAR}}(\phi)
    \right|_{\phi=\phi_n}.
    \label{eq:adapter-update}
\end{equation}
\subsection{Training and Inference}
\label{sec:training-inference}

\paragraph{Training.}
For each training query \(x^{(n)}\), EAR retrieves \(K\) candidates,
performs \(T\) exploration rounds, obtains per-memory feedback, stores
the resulting experience in \(\mathcal{B}_n\), and updates the adapter
using Equation~\eqref{eq:adapter-update}.

\paragraph{Adapter-only inference.}
When the Explorer is disabled at inference time, the final memory set is
selected directly using the adapted relevance logits, 
$    \mathcal{S}^{\mathrm{adapter}}
    =
    \operatorname*{TopS}_{i\in\mathcal{C}}
    z_i^{\phi}.$
This setting does not require critic calls and evaluates the knowledge
assimilated into the adapter.

\paragraph{Explorer-assisted inference.}
When the Explorer is enabled, EAR performs \(T\) exploration rounds
and returns $    \mathcal{S}^{\mathrm{EAR}}
    =
    \mathcal{S}^{(T)}.$
This setting provides additional query-specific adaptation at the cost
of \(T-1\) additional critic calls relative to single-pass retrieval.

\section{Experiments}
\label{sec:exp-setup}
\subsection{Implementation Details}
We equip the EAR framework with the following three retrievers, namely, Contriever \citep{contriever}, Stella  \citep{stella} and GTE \citep{gte}. These retrievers have strong semantic representation capabilities and are widely used in personalized dialogue systems 
For generating answers, we use GPT 5 Mini \citep{singh2025openaigpt5card}. For the explorer, we use $T = 4$ exploration rounds (this choice is explained in \textbf{Appendix \ref{app:hyperparamters}}).
\subsection{Datasets Used}
We evaluate our framework on \textsc{LongMemEval} \citep{longmemeval}  and \textsc{LoCoMo} \citep{maharana2024evaluating}. Both of these operate in a chat-based environment, well equipped for evaluating long-term memory capabilities of agents. In both the benchmarks, we use dialogues as memory candidates. \\
\noindent \textbf{LongMemEval:} We use the \textsc{LongMemEval}\textsubscript{S} version of the benchmark that contains 500 questions, each tied to user-assistant chat histories.\\
\noindent \textbf{LoCoMo:} We use \textsc{LoCoMo}, a conversational benchmark containing 10 conversations.  It encompasses 1540 queries divided into 4 distinct categories such as Single-hop , Multi-hop, Temporal and Open-Domain (we ignore Adversarial queries due to lack of availability of gold-memories).
Table \ref{tab:dataset_stats} details the statistics of each dataset.
\begin{table}[t]
\centering
\small
\begin{tabular}{llr}
\toprule
\textbf{Dataset} & \textbf{Metric} & \textbf{Number} \\
\midrule
\multirow{4}{*}{LoCoMo}
& Queries         & 1,540 \\
& Conversations   & 10 \\
& Sessions        & 272 \\
& Dialogue Turns  & 5,882 \\
\midrule
\multirow{3}{*}{LongMemEval}
& Queries         & 500 \\
& Sessions        & 19,195 \\
& Dialogue Turns  & 199,509 \\
\bottomrule
\end{tabular}
\caption{Statistics of the datasets used.}
\label{tab:dataset_stats}
\end{table}
\subsection{Baselines}
To evaluate the effectiveness of EAR, we compare it against the following baselines.\\
\noindent \textbf{Only Retriever}: We use retriever to select the top-$s$ memories $\mathcal{S}$ based on semantic similarity alone.  \\
\noindent \textbf{Reranker + RL update}: Reranker + RL update proposed by \cite{rmm}.
\subsection{Training and Evaluation}
\label{subsec:training-and-eval}
We conduct our experiments in two phases. In the training phase, we use 80\% queries where the adapter is continually trained. The other 20\% queries are used for evaluation. To evaluate the efficiency of our retrieval mechanism, we restrict the target memory subset size to $s = 5$. This strict capacity constraint establishes a rigorous baseline, demonstrating that EAR can successfully surface critical evidence even when operating under a heavily restricted context budget. We use \emph{Recall@5} (all gold memories in top-5) and \emph{Hitrate@5} (any gold memory in top-5; \citep{ibm_watsonx_hit_rate}) to evaluate retrieval quality. \emph{F1} and \emph{LLM-As-a-Judge} (denoted by \emph{J}) measure question-answering accuracy. We employ turn-level granularity. Unless specified, we use Explorer-assisted inference for the experiments.
\noindent \textbf{Citation Scores.}
To ensure reproducible evaluation devoid of LLM variance, we use a probabilistic reward simulator calibrated to real-world LLM precision/recall reported by \cite{rmm} (Precision 88.0, Recall 86.0. More details in Appendix \ref{app:citation_scores}) This also allows for systematic analysis of impact of feedback noise.

\begin{table*}[t]
\centering
\small
\setlength{\tabcolsep}{4pt}
\begin{tabular}{ll|cccc|cccc}
\toprule
\multicolumn{2}{c|}{} 
& \multicolumn{4}{c|}{\textbf{LongMemEval}} 
& \multicolumn{4}{c}{\textbf{LoCoMo}} \\

\hline
\textbf{Method} & \textbf{Retriever} 
& \textbf{Recall@5} & \textbf{Hitrate@5} & \textbf{F1} & \textbf{J}
& \textbf{Recall@5} & \textbf{Hitrate@5} & \textbf{F1} & \textbf{J} \\

\hline

\multirow{3}{*}{Only Retriever}
& Contriever & 62.37 & 86.02 & 36.12 & 54.84 & 42.60 & 52.01 & 25.23 & 45.45 \\
& Stella & 66.67 & 90.32 & 38.23 & 54.84 & 54.22 & 63.31 & 29.42 & 57.02 \\
& GTE & 68.89 & 93.55 & 40.98 & 60.22 & 57.59 & 71.42 & 33.38 & 62.72 \\

\hline

\multirow{3}{*}{\makecell{Reranker \\ + RL Update}}
& Contriever& 59.71 & 82.08 & 35.06 & 49.46 & 25.97 & 34.31 & 17.11 & 30.08 \\
& Stella & 54.80 & 80.65 & 33.74 & 50.90 & 51.95 & 66.53 & 31.43 & 58.12 \\
& GTE & 33.11 & 68.10 & 25.44 & 37.99 & 39.31 & 52.27 & 27.87 & 49.89 \\

\hline

\multirow{3}{*}{\textbf{EAR (Ours)}}
& Contriever & \textbf{80.28} & \textbf{94.98} & \textbf{41.34} & \textbf{65.23} & 55.06 & 70.83 & 29.85 & 54.44 \\
& Stella & 73.47 & 93.55 & 40.49 & 62.37 & \textbf{63.73} & \textbf{79.61} & 34.10 & 63.09 \\
& GTE & 62.90 & 91.76 & 38.41 & 58.42 & 59.74 & 77.81 & \textbf{35.02} & \textbf{64.94} \\

\hline
\end{tabular}
\caption{Retrieval and Question-Answering performance on LongMemEval and LoCoMo. 
\textbf{Recall@5} and \textbf{Hitrate@5} denote retrieval quality metrics. \textbf{F1} denotes F1-Score between LLM-generated answer and gold answer. \textbf{J} denotes the question-answering accuracy using LLM-as-a-Judge. Results are averaged over 3 runs.}
\label{tab:main_results}
\end{table*}
\subsection{Results and Analysis}

This section first demonstrates overall performance of our EAR framework over the baselines on LongMemEval benchmark. 
We then conduct various analysis including the sample efficiency and feedback-noise robustness.

\subsubsection{Performance Comparison}
On both the benchmarks, we show that our method consistently outperforms the baselines (Table \ref{tab:main_results}). On the LongMemEval dataset, our EAR framework achieves significant improvement in retrieval performance, showing an absolute  \textbf{increase in Recall@5 of 20.6\%} over the existing RL update algorithm and \textbf{17.9\% }over the baseline retriever with Contriever. This is consistent with the LoCoMo dataset too (\textbf{29.1\%} over baseline RL update and \textbf{12.5\%} over baseline retriever).
The gains in retrieval quality translate to downstream question-answering accuracy, improving \textbf{LLM-As-a-Judge score by 10.4\% }(LongMemEval) and \textbf{9.0\%} (LoCoMo) over baseline retriever (Contriever); and \textbf{15.8\%} (LongMemEval) and \textbf{24.4\%} (LoCoMo) over baseline RL update. The baseline RL update notably leads to a performance drop below the static retriever. We discuss this in next section.

With more training queries in LoCoMo, we see that EAR with larger retrievers like Stella and GTE outperform EAR with Contriever. But with fewer training queries in LongMemEval, EAR with Contriever outperforms larger retrievers. This suggests that under tight sample constraints EAR gives the largest improvement for smaller retrievers such as Contriever. This is likely due to number of samples required for training a larger adapter (size scales as $2\times d_{embed}^2$ for more details, see \textbf{Appendix \ref{app:impact-retriever}}). 
 


EAR is model-agnostic and plugs into standard retrieval pipelines. To test compatibility, we layer EAR on top of a fine-tuned text reranker and observe consistent gains over the reranker alone (details and full tables in \textbf{Appendix~\ref{sec: non-rl-rerankers}}). This indicates EAR complements strong non-RL rerankers rather than merely replacing them.



\begin{figure}[t]
    \centering
    \includegraphics[width=0.65\linewidth]{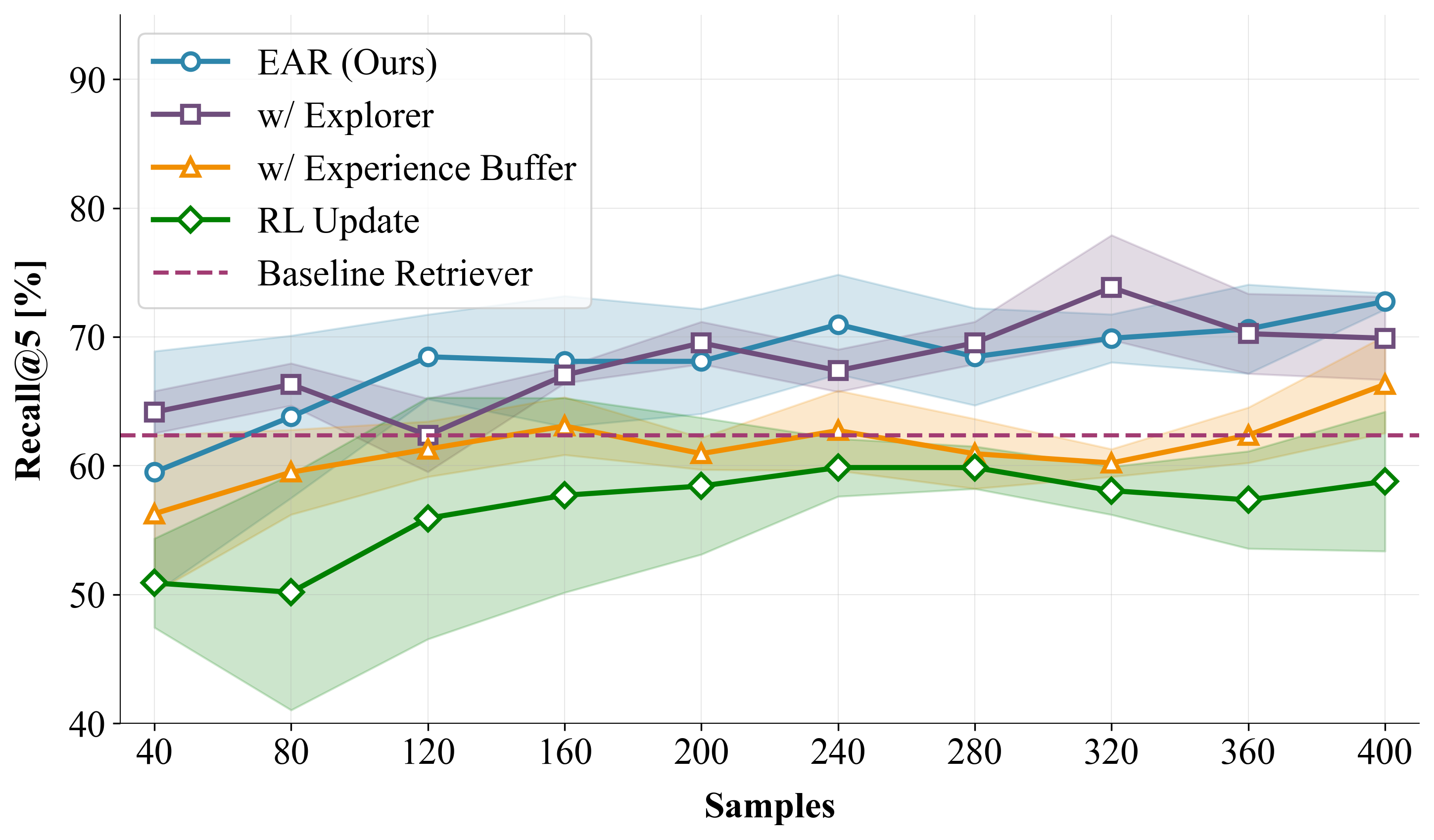}
    \caption{Sample efficiency comparison on LongMemEval (Explorer is disabled at test time). Each curve represents $\text{Mean} \pm \text{Std}$ for 3 runs.}
    \label{fig:efficiency}
\end{figure}

\begin{figure}[t]
    \centering
    \includegraphics[width=0.65\linewidth]{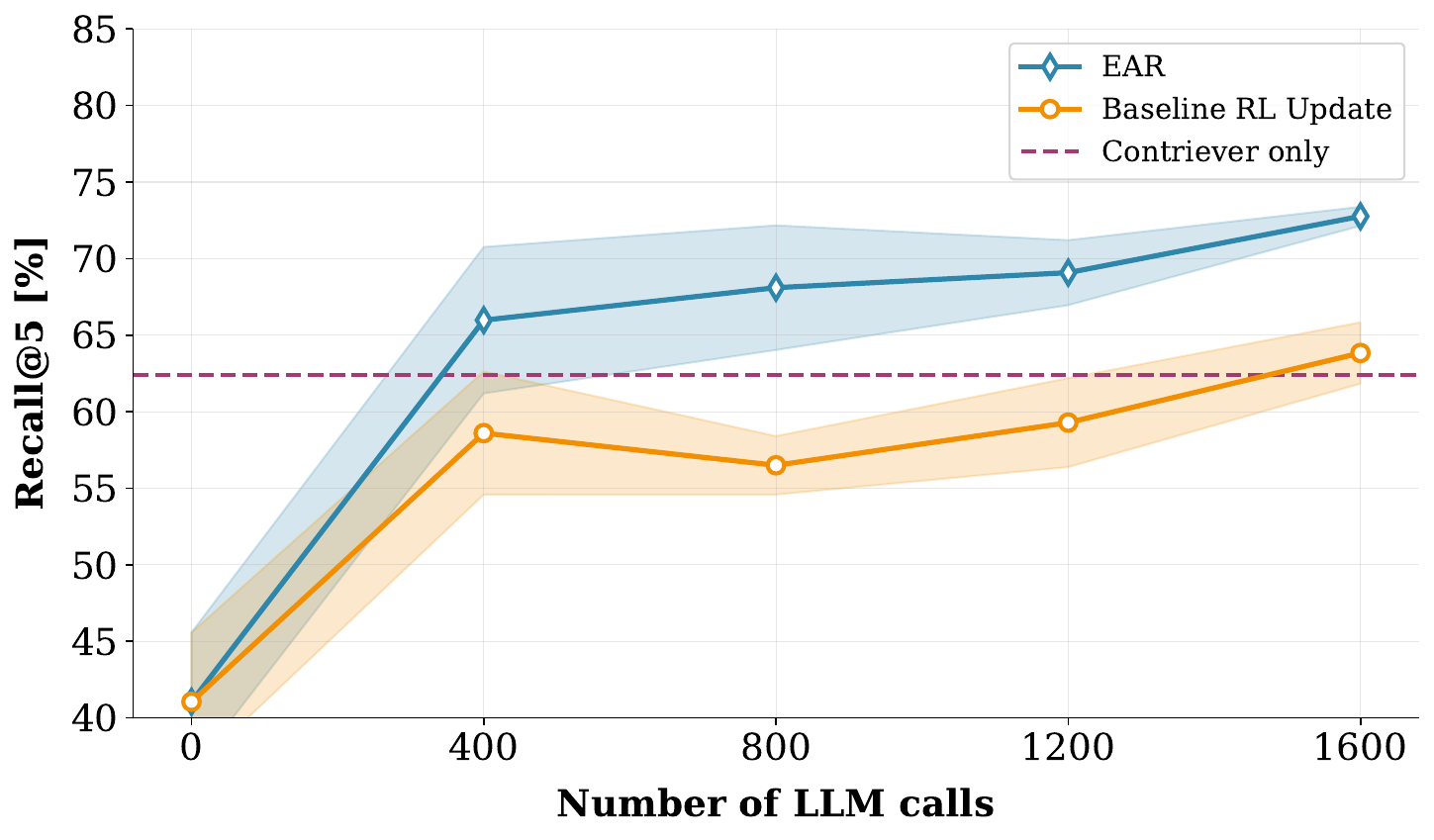}
    \caption{Comparison with Baseline RL update under same computational budget (Explorer is disabled at test time). Each curve represents $\text{Mean} \pm \text{ Std}$ for 3 runs.}
    \label{fig:llm_calls}
\end{figure}

\subsubsection{Sample Efficiency}
Figure~\ref{fig:efficiency} compares the sample efficiency of EAR against the conventional RL update \citep{rmm} on LongMemEval. We disable the explorer at test time to isolate the learned adapter's performance. Comparing efficiency by the number of unique training queries observed, EAR achieves the strongest performance throughout the learning trajectory. While baseline RL update drops 10\% below the static retriever initially, \textbf{EAR limits this degradation to 3\%} and \textbf{surpasses the static retriever after only 80 training queries}. 

In our low-resource continual adaptation setting (400 unique queries), the conventional RL method underperforms the static baseline due to sample inefficiency. To ensure a fair comparison of computational overhead (Figure \ref{fig:llm_calls}), we also evaluate performance normalized by the total number of LLM calls. When the oracle budget is matched ($400 \times T \text{ Exploration Steps} = 1600$ LLM calls; for baseline this was done by running the 400 query-train set for 4 epochs), the baseline RL update overtakes the static retriever (after 1400 LLM calls vs EAR's 320). But even when controlling for this feedback budget, EAR maintains a stable lead.

Finally, Figure~\ref{fig:efficiency} shows this improvement arises from EAR’s complementary components. The Explorer mitigates mainly the severe cold-start problem, stabilizing learning under limited samples. The Experience Buffer then accelerates the overall learning trajectory, enabling the model to quickly outperform the baseline. Together, they resolve the key weaknesses of poor initial performance and inefficient learning of the prior methods.

\begin{figure}[t]
    \centering
    \includegraphics[width=0.65\linewidth]{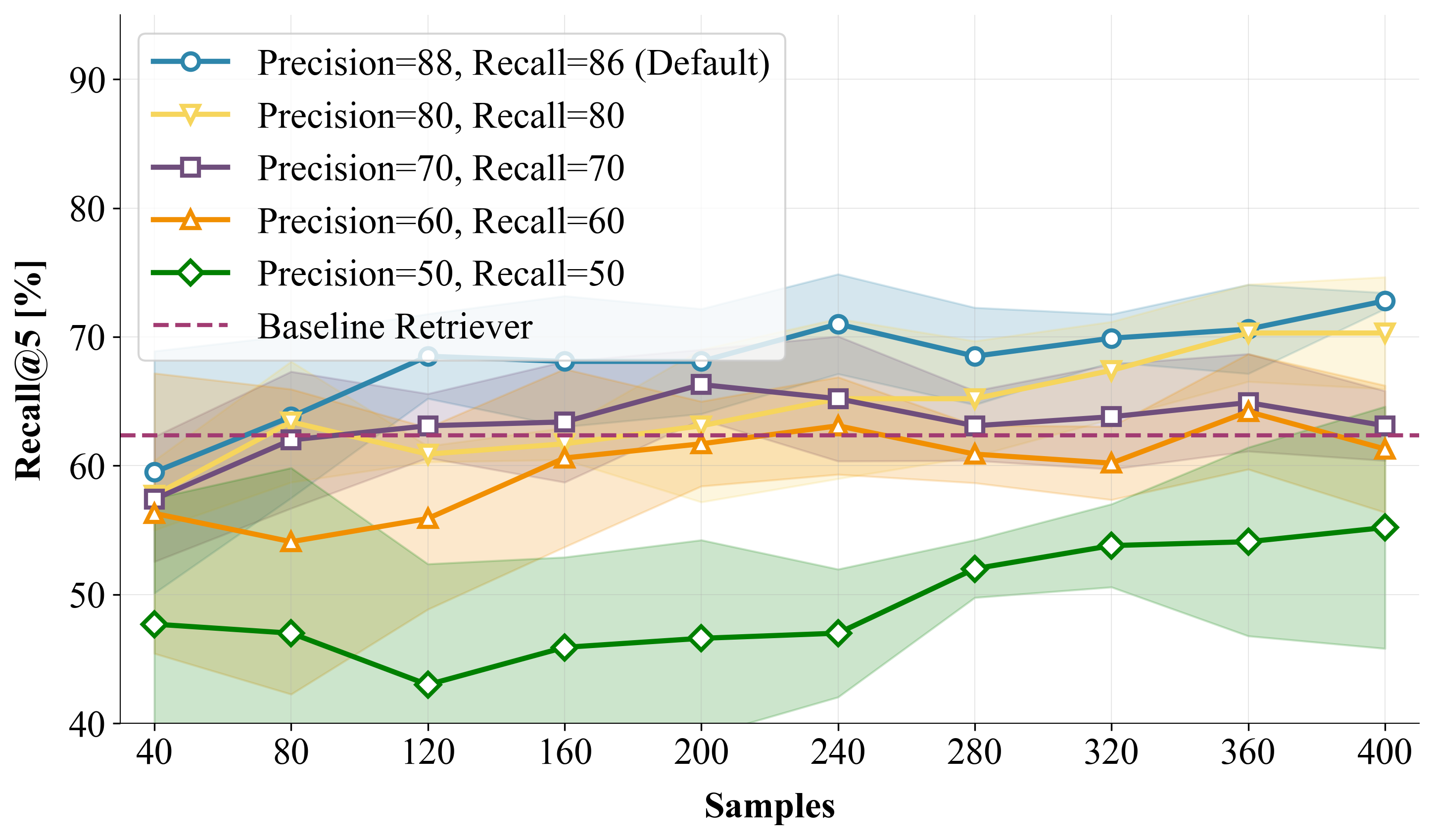}
    \caption{EAR's robustness against feedback noise (Explorer is
disabled at test time). Each curve represents $\text{Mean} \pm \text{ Std}$ for 3 runs. }
    \label{fig:p-r-comparison}
\end{figure}


\subsubsection{Impact of Feedback Quality}
As discussed in Section \ref{subsec:training-and-eval}, we simulate the LLM's citation score using a probabilistic model. Real-world LLMs are prone to distinct biases and hallucinations which manifest as degraded precision and recall in citation scoring. To investigate the effect of feedback quality, we conduct experiments with varying precision and recall. 
Figure \ref{fig:p-r-comparison} indicates how lower feedback quality from the LLM affects the adapter's update process. By averaging the feedback from multiple trials generated by the Explorer, EAR stays resilient to high feedback noise, maintaining a stable learning curve. Even in highly noisy conditions (Precision = 70, Recall = 70) EAR outperforms baseline retriever. To test validity of simulations, we test with real LLM citation scores and found the performance was close to simulations. (See \textbf{Appendix \ref{app:real-world-llm}})

\begin{table}[t]
  \centering
  \small
  \setlength{\tabcolsep}{6pt}
  \renewcommand{\arraystretch}{1}
  \begin{tabular}{lcc}
    \toprule
    \textbf{Variant} & \textbf{LongMemEval} & \textbf{LoCoMo} \\
                     & \textbf{Recall@5}    & \textbf{Recall@5} \\
    \midrule
    \multicolumn{3}{c}{\textit{Without Explorer at test time}} \\
    \midrule
    Adapter \citep{rmm} & 54.12 & 25.97 \\
    + Only Explorer during training & 69.91 & 39.99 \\
    + Only Experience Buffer during training & 66.32 & 29.54 \\
    EAR w/o Explorer at test time & 72.80 & 43.56 \\
    \midrule
    \multicolumn{3}{c}{\textit{With Explorer at test time}} \\
    \midrule
    Only Explorer at test time & 68.81 & 41.20 \\
    \textbf{EAR (Ours)} & \textbf{80.28} & \textbf{55.06} \\
    \bottomrule
  \end{tabular}%
  \caption{\textbf{Ablation study.} The top block reports retrieval performance without using Explorer at test time, isolating the effect of training-time components. The bottom block reports when Explorer is enabled at test time.}
  \label{tbl:ablation}
\end{table}

\subsubsection{Ablation Study}
We conduct an ablation study to isolate the contribution of each component in EAR. Without Explorer at test time, adding either exploration-based training or the Experience Buffer independently already improves retrieval, showing that both components help train a stronger adapter. Combining them yields the best performance, indicating that they contribute complementary training signals. This also shows that EAR allows the costly exploration module to be discarded at
test time, achieving high retrieval accuracy purely through the updated lightweight adapter. When comparing against test-time latency introduced by each method, we see that disabling explorer at test time gives the best balance of latency and performance (see \textbf{Appendix \ref{app:latency}}). 

With Explorer at test time, using Explorer alone without the learned adapter performs worse than using both together. This confirms that EAR performs best when the adapter and Explorer operate synergistically.

\section{Memory Selection from Multiple Stores}
Modern agentic memory systems often have \textit{multiple stores of memory} associated with them, namely, \textit{episodic, semantic} etc. \citep{wang2025mirixmultiagentmemoryllmbased} Every query requires a different proportion of these memory stores to be adequately answered.  In this section we demonstrate EAR's capability to extract the relevant query-dependent mixture of memories.

\subsection{Experimental Setup}

We use the LoCoMo dataset, treating \textit{dialogues as episodic memory} and session-wise \textit{observations as semantic memory}. To isolate the effect of memory mixing \emph{after} retrieval, we enforce a balanced candidate set at the retriever stage: instead of retrieving the top-$k$ memories jointly from the union of both stores, the retriever independently fetches the top-$\frac{k}{2}$ dialogues and the top-$\frac{k}{2}$ observations. The resulting combined set of size $k$ is then passed to the \textit{EAR} framework, from which the final top-$s$ memories are selected ($k=20$, $s=5$). This design prevents the initial candidate pool from being dominated by either observations or dialogues. For the \textit{Retriever Only} baseline, the final top-$s$ memories are selected directly from this same combined candidate set. We use Contriever as the base retriever.
\subsection{Results and Analysis}
This section demonstrates how EAR performs against baseline retriever when presented with multiple memory types (episodic and semantic). 

\begin{table}[t]
\centering
\setlength{\tabcolsep}{4pt}
\renewcommand{\arraystretch}{1.1}

\begin{tabular}{llcc}
\toprule
\textbf{Method} & \textbf{Type} & \textbf{F1} & \textbf{J} \\
\midrule

\multirow{3}{*}{\textbf{Retriever Only}} & Only Dialogues          &  25.23 & 45.45 \\
                                        & Only Observations       &  23.77 & 34.74 \\
                                        & Dialogue + Observations &  28.88 & 47.73  \\
\midrule

\multirow{3}{*}{\textbf{EAR}}           & Only Dialogues          & 31.27 & 54.22\\
                                        & Only Observations       & 25.83 & 39.94\\
                                        & Dialogue + Observations & \textbf{34.40} & \textbf{59.09} \\
\bottomrule
\end{tabular}%

\caption{Results across different input types in LoCoMo dataset. \textbf{J denotes accuracy with LLM-as-a-Judge}}
\label{tab:results_input_type}
\end{table}

\begin{table}[t]
\centering

\setlength{\tabcolsep}{4pt}
\renewcommand{\arraystretch}{1.1}

\begin{tabular}{llccc}
\toprule
\textbf{Method} & \textbf{Type} & \makecell{\textbf{Recall@5} \\ \textbf{Dialogue}} & \makecell{\textbf{Recall@5} \\ \textbf{Observations}} & \textbf{J} \\
\midrule

\multirow{4}{*}{\textbf{Retriever Only}}     & Single Hop    & 32.95 & 63.58 & 61.85 \\          

                                        & Multi Hop     & 1.89 & 16.98 & 20.75  \\  
                                        & Temporal      & 26.56 & 60.94 & 37.50 \\
                                        & Open Domain   & 5.56 & 22.22 & 27.78  \\
\midrule

\multirow{4}{*}{\textbf{EAR}}           & Single Hop      & 54.33 & 62.43 & 73.99 \\
                                        & Multi Hop      & 7.55 & 16.67 & 30.19 \\
                                        & Temporal         & 46.88 & 53.13 & 51.56 \\
                                        & Open Domain     & 16.67 & 22.22 & 27.78 \\
\bottomrule
\end{tabular}%

\caption{Results across query types in LoCoMo dataset. Both dialogues and observations are retrieved. \textbf{J denotes the accuracy with LLM-as-a-Judge}}
\label{tab:results_question_type}
\end{table}

\begin{figure}[t]
    \centering
    \includegraphics[width=0.75\linewidth]{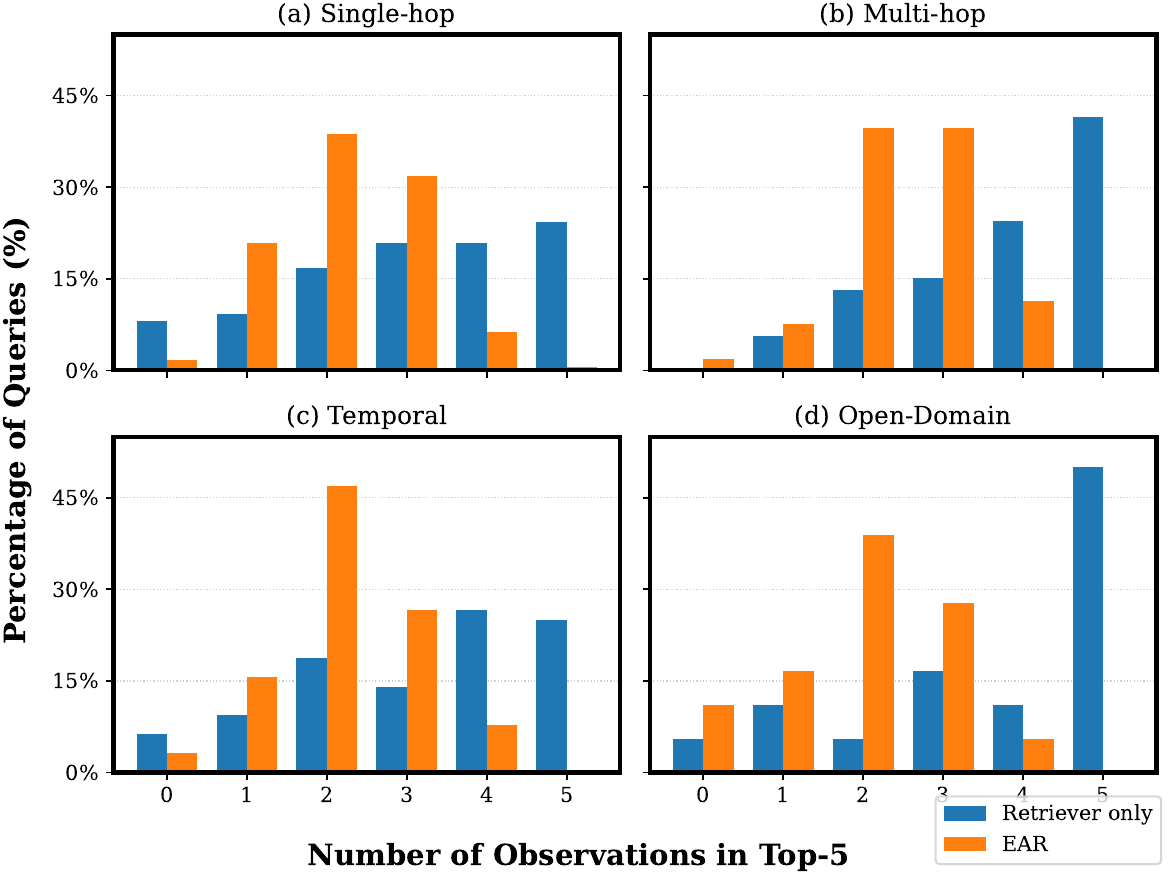}
    \caption{Number of observations chosen in Top-5 across different question types. We see that retriever almost always leans towards picking more observations in its top-5 whereas EAR's top-5 is more balanced.}
    \label{fig:num_observations}
\end{figure}

From Table \ref{tab:results_input_type}, we observe that providing both dialogues and observations in context benefits both methods (EAR and Retriever only). However, EAR makes substantially better use of this mixed memory pool. Figure \ref{fig:num_observations} shows that \textit{Retriever only} selects mostly observations in top-$s$, likely because observations contain denser semantic information and achieve higher similarity scores with the query. \textbf{In contrast, EAR uses feedback to adapt this balance, benefitting from both types of memory.}

From Table~\ref{tab:results_question_type}, compared to the retriever baseline, EAR consistently improves \textbf{Recall@5} for dialogues across all query types and maintains \textbf{Recall@5} for observations. This suggests that EAR selects more of the required dialogues in its top-$5$, which in turn improves downstream accuracy. For example, for \emph{Temporal} queries, \textbf{J} increases from \textbf{37.50 to 51.56}. This indicates having an observation-dominant top-$5$ is insufficient for answering \emph{Temporal} queries. EAR instead retrieves more of the relevant dialogues into the top-$5$, which substantially improves \textbf{J}. For \emph{Open Domain} queries, retrieval improves, but answering accuracy remains unchanged, suggesting that performance there depends more on the LLM’s parametric knowledge than on memory selection.

\paragraph{Geometric Interpretation.}
We use a controlled diagnostic setup to study the embedding space. We curate a balanced episodic/semantic memory pool and store-specific queries, and measure how the adapter changes the query-store proximity in embedding space. After applying the trained adapter, episodic memory dependent queries move closer to episodic memories and semantic memory dependent queries move closer to semantic memories (\textbf{Appendix~\ref{app:qualitative_analysis}, Tables~\ref{tab:qualitative_queries}--\ref{tab:qualitative_queries_dist}}).

\section{Conclusion}
In conclusion, we introduced the Exploratory-Assimilating Reflection (EAR) an adaptive memory retrieval framework designed to improve both retrieval quality and sample-efficient adaptation. Experiments show that our framework significantly outperforms existing adaptive methods, mitigating the severe sample inefficiency of prior approaches. We also show that EAR can retrieve an effective mixture of heterogenous memory types without any architectural changes. Finally, our analysis demonstrate  EAR's robustness against feedback noise, supporting its use in noisy real world settings.

\section*{Limitations}
While this work demonstrates a promising direction, we acknowledge several limitations that open avenues for future research. First, there is further scope for improving the strategies for utilizing experiences from the buffer; developing more sophisticated sampling or replay mechanisms is a key area for future work. Second our findings suggest that performance gains diminish as the size of the base retriever is increased. This warrants a better and more efficient RL algorithm to update the adapter effectively. In future works, we aim to investigate efficient learning techniques such as Low-Rank Adaptation (LoRA) for training adapters for larger and stronger retrievers.




\bibliographystyle{unsrtnat}
\bibliography{ref}
\newpage

\appendix
\section{Implementation Details}
\label{app:implementation}
For ease of reproduction, we provide the exact algorithms for the individual modules along with the hyperparameters used.

\begin{algorithm}[h]
\small
\caption{Exploratory Reflection Algorithm}
\label{alg:short_term_reflection}

\begin{algorithmic}[1]
\item[]\textbf{Input}: Query $\widetilde{\mathbf{q}}$, memory candidates $\widetilde{\mathcal{C}} = \{\widetilde{\mathbf{m}}_1,\dots,\widetilde{\mathbf{m}}_K\}$\\
\item[] \textbf{Output}: Selected memory set $\mathcal{S}$
    \STATE Initialize $\mathcal{S} \gets \emptyset$
    \STATE Initialize $U_i \gets 0$ for all
    $i \in \{1,\dots,K\}$
    \STATE Initialize empirical mean rewards
    $\widehat{\mu}_i \gets 0$, pull counts $N_i \gets 0$,
    and total pulls $N \gets 0$
    \STATE Initialize candidate pool
    $\mathcal{R} \gets \widetilde{\mathcal{C}}$
    \FOR{$j = 1$ to $s$}
        \FOR{each $\widetilde{\mathbf{m}}_i \in \mathcal{R}$}
            \STATE
            $z_i
            \gets
            \operatorname{Sim}
            (\widetilde{\mathbf{q}},
            \widetilde{\mathbf{m}}_i)$
            \hfill \textit{// query--memory similarity}
            \STATE
            $D_i
            \gets
            \max\limits_{\widetilde{\mathbf{m}}_\ell \in \mathcal{S}}
            \operatorname{Sim}
            (\widetilde{\mathbf{m}}_i,
            \widetilde{\mathbf{m}}_\ell)$
            \hfill \textit{// diversity penalty}
            \STATE
            $U_i
            \gets
            \widehat{\mu}_i
            +
            \alpha \cdot
            \sqrt{\tfrac{\log(1+ N)}{(1+ N_i)}}$
            \hfill \textit{// UCB exploration bonus}
        \ENDFOR
        \STATE
        $i^\ast
        \gets
        \arg\max\limits_{\widetilde{\mathbf{m}}_i \in \mathcal{R}}
        \left\{
            \lambda_{\mathrm{rel}} z_i
            -
            \lambda_{\mathrm{div}} D_i
            +
            \lambda_{\mathrm{exp}} U_i
        \right\}$
        \STATE
        $\mathcal{S}
        \gets
        \mathcal{S}
        \,\|\,
        \widetilde{\mathbf{m}}_{i^\ast}$
        \hfill \textit{// append chosen memory}
        \STATE
        $\mathcal{R}
        \gets
        \mathcal{R}
        \setminus
        \{\widetilde{\mathbf{m}}_{i^\ast}\}$
        \hfill \textit{// remove from candidate pool}
    \ENDFOR
    \STATE Obtain LLM citations
    $y_i \in \{-1,1\}$ for each
    $\widetilde{\mathbf{m}}_i \in \mathcal{S}$
    \FOR{each $\widetilde{\mathbf{m}}_i \in \mathcal{S}$}
        \STATE
        $N_i \gets N_i + 1$;
        \;
        $N \gets N + 1$
        \STATE
        $\widehat{\mu}_i
        \leftarrow
        \widehat{\mu}_i
        +
        \frac{
            y_i-\widehat{\mu}_i
        }{
            N_i
        }$
    \ENDFOR
\end{algorithmic}
\end{algorithm}

For Algorithm \ref{alg:short_term_reflection}, we use the following hyperparameters. The hyperparameters for the Explorer were chosen empirically, to achieve right balance between similarity, diversity and exploration. The weights are designed in such a way to give equal importance to memory slate diversity and exploration within the retrieval space and slightly higher importance to similarity to candidate memories. The UCB coeffecient was set to 0.5 to achieve a balance between exploration and exploitation.
\begin{itemize}
    \item Required Subset Size $s$: 5
    \item Coefficients
    $\lambda_{\mathrm{rel}},
    \lambda_{\mathrm{div}},
    \lambda_{\mathrm{exp}}$:
    0.4, 0.3, 0.3
    \item UCB coefficient $\alpha$: 0.5
\end{itemize}

\paragraph{Step-by-step walkthrough}

\begin{itemize}
    \item The explorer recieves the query and memory embeddings as input. In the EAR setup, the query and memory embeddings are modified beforehand by the adapter.
    \item (Line 2-13) Initially the required subset $\mathcal{S}$ empty. Then we calculate similarity $z_i$, diversity $D_i$ and UCB $U_i$ for each memory $\widetilde{\mathbf{m}}_i$. The memory maximizing the weighted average of the three scores is chosen. This chosen memory is removed from the candidate pool. This step continues till the desired subset size $|\mathcal{S}| = s$ is achieved.
    \item (Line 15) Once the final subset is formed, an LLM-evaluator provides binary citations (+1, -1) for each memory indicating its usefulness for answering the query.
    \item (Line 16 - 18) After this we update the UCB statistics (for each memory $\widetilde{\mathbf{m}}_i$ - number of times it was chosen $N_i$, mean reward $\widehat{\mu}_i$ and total number of arm pulls done across all memories, $N$).
    \item The above steps constitute one iteration of exploration.
\end{itemize}

\subsection{Assimilating reflection}
Algorithm \ref{alg:ear_replay} gives the step-by-step implementation for the Assimilating Reflection component. Following \citep{rmm} for our experiments, we use the following set of hyperparameters:

\begin{algorithm}[t]
\small
\caption{Assimilating Reflection with Experience Replay}
\label{alg:ear_replay}

\begin{algorithmic}[1]
\item[]\textbf{Input}: Query $\mathbf{q}^{(n)}$, memory candidates
$\mathcal{C}^{(n)}$= $\{\mathbf{m}_1^{(n)},\dots,\mathbf{m}_K^{(n)}\}$
, replay buffer $\mathcal{B}_{n-1}$\\
\item[] \textbf{Output}: Updated $\phi=\{W_q,W_m\}$

\STATE Adapt embeddings:
$\widetilde{\mathbf{q}}^{(n)}
\gets
\mathbf{q}^{(n)}+W_q\mathbf{q}^{(n)}$;\;
$\widetilde{\mathcal{C}}^{(n)}
\gets
\{
\widetilde{\mathbf{m}}_i^{(n)}
=
\mathbf{m}_i^{(n)}
+
W_m\mathbf{m}_i^{(n)}
\}_{i=1}^{K}$
\STATE Loss
$\mathcal{L}_{\mathrm{cur}}^{(n)}(\phi)\gets 0$
\FOR{$t=1$ to $T$}
  \STATE
  $\mathcal{S}^{(n,t)}
  \gets
  \textsc{ExploratoryReflection}
  (
      \widetilde{\mathbf{q}}^{(n)},
      \widetilde{\mathcal{C}}^{(n)},
      T
  )$
  \STATE
  $\mathbf{y}^{(n,t)}
  \gets
  \textsc{Citations}
  (\mathcal{S}^{(n,t)})$
  \hfill \textit{// LLM citation score}
  \STATE Update UCB statistics using
  $\mathcal{S}^{(n,t)}$ and
  $\mathbf{y}^{(n,t)}$
  \STATE
  $\mathcal{L}_{\mathrm{cur}}^{(n)}(\phi)
  \gets
  \mathcal{L}_{\mathrm{cur}}^{(n)}(\phi)
  -
  \frac{1}{|\mathcal{S}^{(n,t)}|}
  \sum_{i\in\mathcal{S}^{(n,t)}}
  \left(
      y_i^{(n,t)}-b
  \right)
  \log
  p_\phi
  \left(
      i
      \mid
      \mathbf{q}^{(n)},
      \mathcal{C}^{(n)}
  \right),$
\ENDFOR
\STATE
$\mathcal{L}_{\mathrm{cur}}^{(n)}(\phi)
\gets
\mathcal{L}_{\mathrm{cur}}^{(n)}(\phi)/T$
\STATE Store per-memory citations over $T$ explorations in
$\overline{\mathbf{y}}^{(n)}$
\STATE
\textbf{Store experience}:
\STATE
$\mathcal{B}_{n}
\gets
\mathcal{B}_{n-1}
\cup
\{
(\mathbf{q}^{(n)},\mathcal{C}^{(n)},
\overline{\mathbf{y}}^{(n)})
\}$
\vspace{0.25em}
\STATE \textbf{Replay and update}
\STATE Sample mini-batch
$\mathcal{J}_n
\gets
\operatorname*{TopB}_{r<n}
\operatorname{Sim}
\left(
    \mathbf{q}^{(n)},
    \mathbf{q}^{(r)}
\right)$
\STATE Experience replay loss
$\mathcal{L}_{\mathrm{rep}}^{(n)}(\phi)\gets 0$
\FOR{$r=1$ to $|\mathcal{J}_n|$}

    \STATE Adapt embeddings
    $\mathbf{q}^{(r)}$ and
    $\mathcal{C}^{(r)}$
  \STATE Sample
  $\widetilde{\mathcal{S}}^{(r)}$
  using similarity-scores between
  $\widetilde{\mathbf{q}}^{(r)}$ and
  $\widetilde{\mathcal{C}}^{(r)}$
  with Gumbel trick
  \STATE Recompute reward
  $\widetilde{\mathbf{r}}^{(r)}$
  for
  $\widetilde{\mathcal{S}}^{(r)}$
  using
  $\overline{\mathbf{y}}^{(r)}$
  \STATE
  $\mathcal{L}_{\mathrm{rep}}^{(n)}(\phi)
  \gets
  \mathcal{L}_{\mathrm{rep}}^{(n)}(\phi)
  -
  \frac{1}{|\widetilde{\mathcal{S}}^{(r)}|}
  \sum_{i\in\widetilde{\mathcal{S}}^{(r)}}
  \left(
      \widetilde{r}_i^{(r)}-b
  \right)
  \log
  p_\phi
  \left(
      i
      \mid
      \mathbf{q}^{(r)},
      \mathcal{C}^{(r)}
  \right),$
\ENDFOR
\STATE
$\mathcal{L}_{\mathrm{rep}}^{(n)}(\phi)
\gets
\mathcal{L}_{\mathrm{rep}}^{(n)}(\phi)
/|\mathcal{J}_n|$
\STATE
$\phi
\gets
\phi
-
\eta
\nabla_\phi
\left(
    \mathcal{L}_{\mathrm{cur}}^{(n)}(\phi)
    +
    \lambda_{rep}\mathcal{L}_{\mathrm{rep}}^{(n)}(\phi)
\right)$

\end{algorithmic}
\end{algorithm}

\begin{itemize}
    \item Batch size: 4
    \item Learning Rate: 0.001
    \item Baseline $b$: 0.5 (to give stronger negative penalty for retrieving irrelevant memories)
    \item Gumbel Temperature: 0.5
    \item Experience replay batch size: 4
    \item Experience replay strength $\lambda_{rep}$: 1.0 
\end{itemize}

\paragraph{Step-by-step walkthrough}
\begin{itemize}
    \item We take query $\mathbf{q}^{(n)}$ and top-$K$ memory candidates $\mathcal{C}^{(n)}$ provided by retriever as input.
    \item (Line 1) Initally the replay buffer is empty.
    \item (Line 2) Using the residual linear adapter, the query and memory embeddings are modified.
    \item (Line 3 - 9) Using the modified embeddings, we perform $T$ exploration steps. At each step, the Exploratory Reflection module is invoked to provide a subset of $s$ memories. The LLM evaluator provides citation scores which in turn is used to update UCB statistics. We then calculate the loss.
    \item (Line 11-12) After $T$ explorations, the average citation recieved by each candidate memory is stored in $\overline{\mathbf{y}}^{(n)}$. The query, candidate memories and average citations per memory are stored in the Experience Buffer.
    \item (Line 14) To replay experiences from the Experience Buffer, we sample a mini batch $\mathcal{J}_n$. We sample them based on cosine similarity to the current query $\mathbf{q}^{(n)}$. Note that experience buffer contains the original query and memory embeddings before applying the adapter. Impact of different sampling strategies is discussed in Appendix \ref{app:sampling}.
    \item (Line 17 - 18) For each sampled past experience, we adapt the past query and memory embeddings using the current adapter. To enable stochastic sampling for each past experience we employ the Gumbel-trick. Instead of top-$s$ directly through the updated query and memory embedding, we add Gumbel noise to the similarity scores.
    \begin{equation}
        \widetilde{z}_i^{(r)}
        =
        {\widetilde{\mathbf{q}}^{(r)}}\cdot
        \widetilde{\mathbf{m}}_i^{(r)}
        +
        g_i^{(r)}
    \end{equation}
    Where $g_i^{(r)}
        =
        -\log
        \left(
            -\log
            \left(
                u_i^{(r)}
            \right)
        \right)$ and 
    $u_i^{(r)}
    \sim
    \operatorname{Uniform}(0,1)$.
    Gumbel trick perturbs the similarity scores using uniform noise. The perturbed scores are normalized using a softmax function. After sampling, we get a slate $\widetilde{\mathcal{S}}^{(r)}$.
    \item (Line 19) Since we have average citations for each memory in the experience buffer, we convert it to binary rewards as
    \begin{equation}
        \widetilde{r}_i^{(r)}
        =
        \begin{cases}
            +1,
            & \text{if }
            \overline{y}_i^{(r)}>0,
            \\
            -1,
            & \text{if }
            \overline{y}_i^{(r)}\leq0.
        \end{cases}
    \end{equation}
    Using the calculated reward, loss is calculated.
    \item (Line 23) Using a combination of the current query loss and experience replay loss, optimization is done.
\end{itemize}
\subsection{Simulating Citation Scores}
\label{app:citation_scores}
To eliminate the confounding effects of the LLM on our experiment, we use a probabilistic reward model that labels the selected memories as useful/not useful to answering the query. This replaces the LLM based citation scores in our experiment. Here citations refer to which of the selected top-5 memories where found useful. So the prediction space is limited to $\mathcal{U}$.
\[
\mathcal{U} = \{0,1,2,3,4\}.
\]
Let $\mathcal{G}$ denote the set of unique ground-truth citations, and let $n = |\mathcal{G}|$ be the number of distinct gold citations.

\paragraph{True positive generation.}
To match a target recall $r \in [0,1]$, each ground-truth citation $g \in \mathcal{G}$ is independently included in the prediction set with probability $r$. Therefore, the expected number of true positives is
\[
\mathbb{E}[\mathrm{TP}] = n r.
\]
This directly aligns the expected recall with the target value, since
\[
\mathbb{E}[\mathrm{Recall}] \approx \frac{\mathbb{E}[\mathrm{TP}]}{n} = r.
\]

\paragraph{False positive generation.}
False positives are drawn from the remaining citation indices,
\[
\mathcal{F} = \mathcal{U} \setminus \mathcal{G},
\]
with size
\[
u = |\mathcal{F}| = 5 - n.
\]
Each candidate false positive $f \in \mathcal{F}$ is independently included with probability $p_{\mathrm{fp}}$. This probability is chosen so that the expected precision matches a target value $p \in [0,1]$. Using
\[
\mathrm{Precision} = \frac{\mathrm{TP}}{\mathrm{TP} + \mathrm{FP}},
\]
and substituting the expected counts $\mathbb{E}[\mathrm{TP}] = nr$ and $\mathbb{E}[\mathrm{FP}] = u p_{\mathrm{fp}}$, we solve
\[
p \approx \frac{nr}{nr + u p_{\mathrm{fp}}}.
\]
Rearranging gives
\[
p_{\mathrm{fp}} = \frac{nr(1-p)}{up}.
\]
Thus, the expected number of false positives becomes
\[
\mathbb{E}[\mathrm{FP}] = u p_{\mathrm{fp}},
\]
which yields the desired trade-off between recall and precision.

\section{Additional Results}
\label{sec:additional_results}

\subsection{Impact of Hyperparameters}
\label{app:hyperparamters}
We test our EAR framework on the LongMemEval dataset with different hyper-parameters.
The results in Table \ref{tbl:parameters} suggest a trade-off regarding the number of exploration steps. An insufficient number of steps risks overlooking valuable feedback, whereas an excessive number may increase the vulnerability to noise within the feedback signals.
Similarly, our experiments on the batch size used in Experience Replay indicate that an optimal balance must be struck between prioritizing immediate feedback from the current query and assimilating broader lessons from past experiences stored in the buffer.

\begin{table}[t]
  \centering
  \small
  \begin{tabular}{ccc}
    \hline
    \textbf{\# of Exploration} & \textbf{Replay Batch Size} & \textbf{Recall@5} \\
    \hline
     2   &  4  & 68.5              \\
    4   &  4  & \textbf{72.8}      \\
    8   &  4  & 69.5              \\
    4   &  2  & 71.3              \\
    4   &  8  & 71.0              \\
    \hline
  \end{tabular}
  \caption{Impact of the number of exploration and the size of experience buffer. Recall@5 shows recall rate of updated reranker in percentage (\%) (i.e., Explorer is disabled at test time).}
  \label{tbl:parameters}
\end{table}

\begin{table}[t]
  \centering
  \small
  \begin{tabular}{lc}
    \hline
    \textbf{Sampling Strategy} & \textbf{Recall@5} \\
    \hline
    Similarity         & \textbf{72.8}              \\
    Reward            & 69.9              \\
    Random   & 70.3              \\
    \hline
  \end{tabular}
  \caption{Impact of sampling strategies of Experience Replay. Recall@5 shows recall rate of updated reranker in percentage (\%) (i.e., Explorer is disabled at test time).}
  \label{tbl:sampling}
\end{table}

\subsection{Impact of Sampling Strategy}
\label{app:sampling}
We test three sampling strategies for Experience Replay.
\textit{Similarity} strategy retrieves experiences with top-n most similar queries.
\textit{Reward} strategy uses bottom-n experiences by reward. 
Finally, \textit{Random} strategy uniformly samples $n$ experiences from the buffer. 
The results in Table \ref{tbl:sampling} show retrieval quality with different sampling strategies indicate that leveraging past experiences from similar queries is the most effective strategy. We hypothesize that this is because these relevant experiences help to guide the reranker's update direction more effectively.
\subsection{Performance--Cost Tradeoff}
\label{app:latency}
Though EAR achieves significant improvement in performance over the baseline retriever, it introduces additional computational cost in the form of extra latency and LLM call. In this section, we analyze the tradeoff between performance and latency across all compared methods.
\begin{table}[t]
  \centering
  \small
  \begin{tabular}{lcc}
    \hline
    \textbf{Method} & \textbf{Latency (s)} & \textbf{Recall@5} \\
    \hline
     Retriever Only   &  1.80  & 62.37              \\
     Reranker + RL Update & 2.30 & 54.12 \\
     EAR w/o Explorer & 2.30 & 72.80 \\
     EAR w/ Explorer & 16.88 & 80.28 \\
    \hline
  \end{tabular}
  \caption{Latency introduced by each method per query during test time. Performance is shown on LongMemEval dataset.}
  \label{tbl:latency}
\end{table}

Table~\ref{tbl:latency} reports the average test-time latency per query for each method, along with the corresponding Recall@5 on LongMemEval. In these measurements, we assume that the query has already been embedded and that memory embeddings are precomputed and stored. We use a file based storage system for the embeddings. We further exclude the latency of the downstream answer generation step, so the reported numbers isolate the cost of memory retrieval.

The \textit{Retriever Only} baseline incurs the lowest latency, as it simply returns the top-$k$ nearest memories from the embedding store. Adding the lightweight linear adapter increases latency only marginally, from 1.80s to 2.30s per query. This adapter architecture is identical for \textit{Reranker + RL Update} and \textit{EAR}; therefore, when the explorer is disabled at test time, both methods are functionally equivalent in terms of inference cost. However, despite having the same test-time architecture, \textit{EAR w/o Explorer} achieves substantially higher Recall@5 than \textit{Reranker + RL Update}, indicating that EAR's training procedure learns a much more effective adapter.

Enabling the explorer at test time yields the best retrieval performance, improving Recall@5 from 72.80 to 80.28, an absolute gain of 7.48 points over \textit{EAR w/o Explorer}. This gain, however, comes at the cost of additional latency: test-time inference rises from 2.30s to 16.88s per query. The increase is primarily due to the iterative exploration process, which requires an additional $T-1$ LLM calls for each query.

Overall, these results highlight two practical operating points for EAR. \textit{EAR w/o Explorer} offers a strong performance--efficiency balance, delivering a large improvement over the baseline with only a minor latency increase. In contrast, \textit{EAR w/ Explorer} provides the highest retrieval accuracy, but at a substantially higher inference cost. This makes the explorer particularly suitable for settings where retrieval quality is critical and additional latency can be tolerated.

\subsection{Using Real World LLMs for Citations}
\label{app:real-world-llm}
To validate whether our simulated citation scores agree to LLM based citations, we conduct the experimental setup with GPT 5 Mini as the evaluator instead of the probabilistic model. From Table \ref{tbl:real-world-llm} We find the results with the actual LLM are close to the results with simulated probabilistic model. Results for GPT 5 Mini are reported after one run.

\begin{table}[t]
  \centering
  \begin{tabular}{lccc}
    \hline
    \textbf{Citation Scores} & \textbf{Precision} & \textbf{Recall} &  \textbf{Recall@5} \\
    \hline
    Simulated          & 88 & 86 & 72.8              \\
    GPT 5 Mini         & 78 & 86 & 70.1               \\
    \hline
  \end{tabular}
  \caption{Performance with GPT 5 Mini as the evaluator. Recall@5 shows recall rate of updated reranker (\%) in LongMemEval dataset (i.e., Explorer is disabled at test time).}
  \label{tbl:real-world-llm}
\end{table}

\subsection{Impact of Base Retriever Size}
\label{app:impact-retriever}


We evaluate EAR with three base retrievers: Contriever, Stella-1.5B, and GTE-7B. Across both LongMemEval and LoCoMo, we observe that the absolute improvement obtained by EAR decreases as the retriever size increases. A likely reason is that larger retrievers produce higher-dimensional embeddings (Contriever: 768, Stella: 1024, GTE: 3584), which in turn increase the size of the trainable adapter. In our implementation, the number of adapter parameters scales as $2d_{\text{embed}}^2$, where $d_{\text{embed}}$ is the embedding dimension (this comes from the two projection matrices $W_q$ and $W_m$ mentioned in section \ref{sec:adapter}. This suggests that larger retrievers may require more training examples for EAR to realize substantial gains).

Our results support this interpretation. On LongMemEval, which contains 400 training queries, EAR with Stella improves over the baseline retriever but remains below EAR with Contriever. In contrast, on LoCoMo, which provides 1232 training queries, EAR with Stella substantially outperforms EAR with Contriever. A similar trend is observed for GTE: on LongMemEval, EAR does not surpass the baseline retriever, whereas on LoCoMo it achieves clear gains over the baseline. 

\begin{figure}[t]
    \centering
    \includegraphics[width=0.65\linewidth]{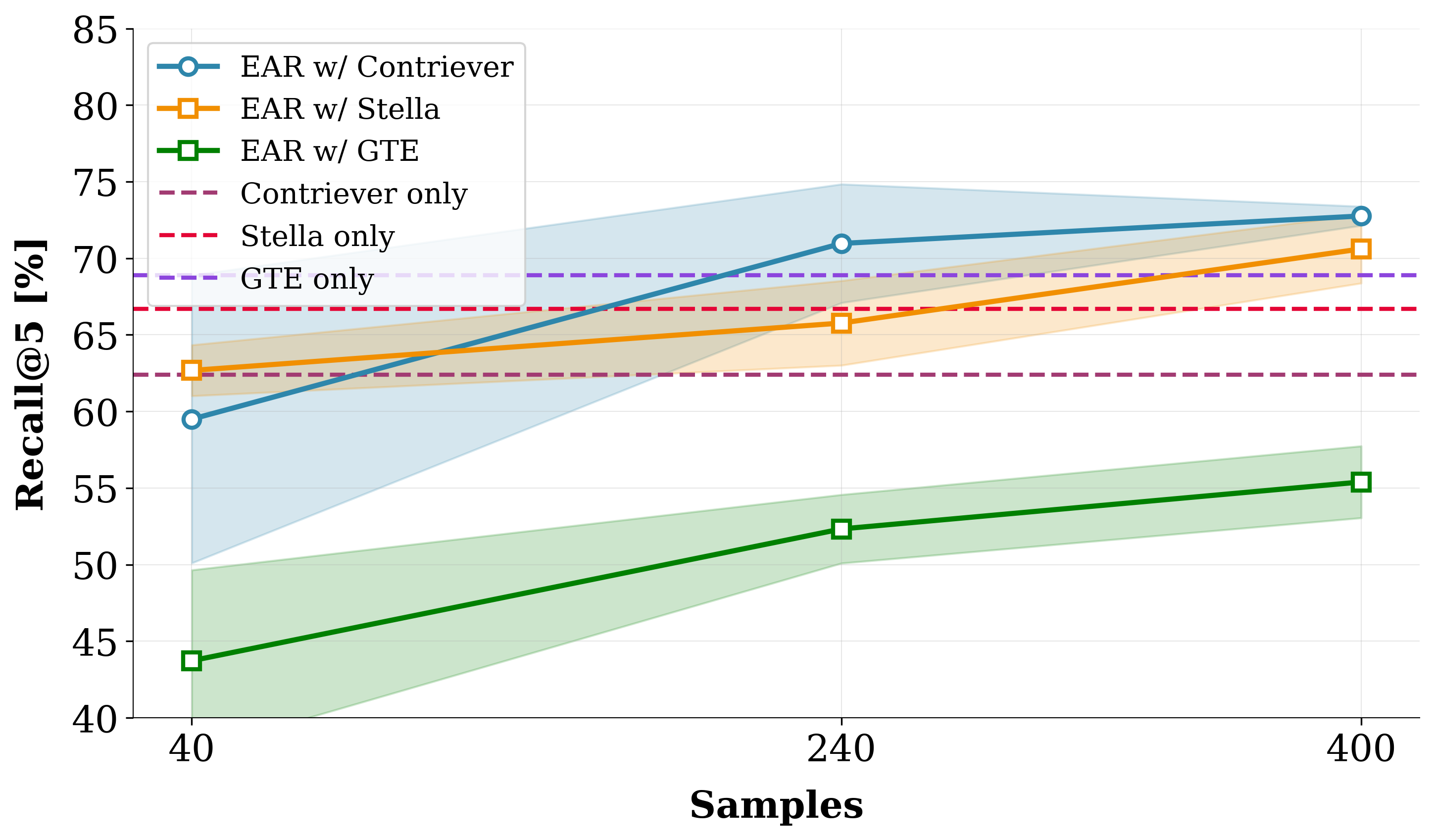}
    \caption{Comparison of recall rates across different retrievers. Recall rate in percentage on LongMemEval. }
    \label{fig:retriever-comparison}
\end{figure}

Figure~\ref{fig:retriever-comparison} further supports this trend on LongMemEval. Although EAR with GTE does not yet outperform the baseline in this low-data regime, its recall improves steadily during training, indicating a positive learning curve similar to the other retrievers. This suggests that with sufficient training data, EAR can also improve over stronger and larger base retrievers.


\subsection{Non-RL based rerankers}
\label{sec: non-rl-rerankers}


In addition to our RL-trained adapter, we evaluate whether EAR can be composed with standard, non-RL rerankers that are trained for generic text reranking. Concretely, we use \textsc{MonoT5-base-MSMARCO}, a T5-base cross-encoder reranker fine-tuned on the MS MARCO passage ranking task. Given a query, \textsc{MonoT5} produces relevance scores for query--memory pairs and reranks the candidate set. EAR is then applied \emph{on top of} this reranked pool: the explorer operates over the same top-$k$ candidates using Contriever embeddings, and we additionally make the \textsc{MonoT5} scores available as side-information during exploration (e.g., as an auxiliary signal when proposing memory subsets).

Table~\ref{tbl:non-rl-rerankers} shows that combining \textsc{MonoT5} with EAR improves \textbf{Recall@5} over using \textsc{MonoT5} alone on both LongMemEval and LoCoMo. This result highlights a key design property of EAR: it is not tied to a particular retrieval stack. Instead, EAR functions as a \emph{retrieval-time, feedback-driven layer} that can be placed downstream of diverse candidate generators and rerankers, leveraging their strengths while correcting failures through iterative selection.

\paragraph{Why feedback-driven exploration matters.}
Static rerankers such as \textsc{MonoT5} are trained to approximate broad, general-purpose relevance notions from large public corpora. In contrast, memory retrieval for agents is often \emph{domain- and objective-dependent} (e.g., code-assistants requiring execution-relevant context, tool-use traces, or repository-specific conventions). In such niche settings, purely static reranking may misalign with the true task objective. EAR's explorer, however, optimizes memory selection using \emph{task feedback} (an external critic signal; an LLM in our implementation), allowing the retrieval process to adapt its behavior to the environment and the downstream success criterion without requiring the reranker to have been pretrained on that niche distribution. We therefore view EAR as complementary to pretrained rerankers: cross-encoders can improve candidate ordering, while EAR provides an adaptive mechanism to refine the final retrieved set under the task's feedback signal.

\begin{table}[h]
  \centering
  \small
  \begin{tabular}{llcc}
    \hline
    \textbf{Dataset} & \textbf{Method} & \textbf{Paramter Size}  & \textbf{Recall@5} \\
    \hline
    \multirow{3}{*}{\textbf{LongMemEval}} & EAR & 1.18M & 80.28 \\
    & MonoT5 & 220M & 82.20              \\
    & MonoT5  + EAR & 221.18M & \textbf{85.50}              \\
    \hline
    \multirow{3}{*}{\textbf{LoCoMo}} & EAR & 1.18M & 55.06 \\
    & MonoT5 & 220M & 61.46              \\
    & MonoT5 + EAR & 221.18M & \textbf{62.08}              \\
    \hline
    
  \end{tabular}
  \caption{Comparison of recall rates when using RL based rerankers with EAR}
  \label{tbl:non-rl-rerankers}
\end{table}

\section{Qualitative Analysis for Selection from Different Memory Stores}
\label{app:qualitative_analysis}
LoCoMo does not provide explicit labels indicating whether a query should be resolved primarily from episodic (event-like) or semantic (fact-like) memories. To probe whether the trained adapter induces \emph{store-consistent} structure in the embedding space, we conduct a controlled diagnostic study with a clearly separated memory pool and a small set of hand-labeled queries.
\begin{table}[h]
  \centering
  \small
  \setlength{\tabcolsep}{4pt}
  \renewcommand{\arraystretch}{1.1}
  \begin{tabular}{clc}
    \toprule
    \textbf{S.No} & \textbf{Query} & \textbf{Memory} \\
    \midrule
    1 & Who took chapter 6 notes on 2025-03-25? & Episodic \\
    2 & When did Kai and Rina have their prep meeting on 25th April, 2025? & Episodic \\
    3 & When did Ethan's coffee machine last malfunction?  & Episodic \\
    4 & Where is the meeting agenda doc shared by Kai? & Episodic \\
    5 & At what time is Leo's party happening on 8th March 2025? & Episodic \\
    6 & What is Caroline's preference when texting? & Semantic \\
    7 & What is Leo's preference for meetings? & Semantic \\
    8 & Why does Leo use music for gatherings? & Semantic \\ 
    9 & What are Kai and Rina's strengths? & Semantic \\
    10 & What is Caroline's view towards LGBTQ? & Semantic \\
    \bottomrule
  \end{tabular}
  \caption{Examples of queries and the memory type it predominantly requires.}
  \label{tab:qualitative_queries}
\end{table}

\paragraph{Curated setup.}
We construct a pool of 50 memories with unambiguous store membership: 25 episodic and 25 semantic. We also curate 10 diagnostic queries, where 5 predominantly require episodic memories and 5 predominantly require semantic memories (Table~\ref{tab:qualitative_queries}). These queries are not used for training; they are only used for qualitative analysis.

\paragraph{Proximity metric.}
Let $q$ be the normalized query embedding and $M_e, M_s$ be the sets of normalized episodic and semantic memory embeddings.
We use cosine distance defined as $d(q,m) = 1 - \mathrm{Sim}(q,m)$, where $\mathrm{Sim}(\cdot,\cdot)$ denotes cosine similarity.

\begin{equation}
    d_e(q) = \frac{1}{|M_e|} \sum_{m \in M_e} \bigl(1 - \mathrm{Sim}(q,m)\bigr),
\end{equation}
\begin{equation}
    d_s(q) = \frac{1}{|M_s|} \sum_{m \in M_s} \bigl(1 - \mathrm{Sim}(q,m)\bigr),
\end{equation}
\begin{equation}
    \Delta(q) = d_e(q) - d_s(q).
\end{equation}

A positive $\Delta$ indicates the query is closer to semantic memories than episodic memories (i.e., $d_s(q) < d_e(q)$),
while a negative $\Delta$ indicates higher affinity to episodic memories (i.e., $d_e(q) < d_s(q)$).
We compute $\Delta_{\text{before}}$ using base embeddings and $\Delta_{\text{after}}$ after applying the trained adapter.

\begin{table}[h]
  \centering
  \small
  \setlength{\tabcolsep}{4pt}
  \renewcommand{\arraystretch}{1.1}
  \begin{tabular}{cccc}
    \toprule
    \textbf{Query} & \textbf{Memory Required} & \textbf{$\Delta_{before}$} & \textbf{$\Delta_{after}$} \\
    \midrule
    1 & Episodic & -0.001 & -0.069 \\
    2 & Episodic & -0.023 & -0.079 \\
    3 & Episodic & -0.033 & -0.101 \\
    4 & Episodic & -0.023 & -0.050 \\
    5 & Episodic & -0.008 & -0.073 \\
    6 & Semantic & -0.001 & 0.149 \\
    7 & Semantic & 0.033 & 0.054 \\
    8 & Semantic & 0.002 & 0.064 \\
    9 & Semantic & -0.006 & 0.022 \\
    10 & Semantic & -0.004 & 0.124 \\
    \bottomrule
  \end{tabular}
  \caption{Comparison of the query's proximity to either of the memory types before and after training the adapter.}
  \label{tab:qualitative_queries_dist}
\end{table}
\paragraph{Observation.}
From Table~\ref{tab:qualitative_queries_dist}, we observe that prior to training the adapter, $\Delta_{\text{before}}$ values are close to zero, indicating weak separation between episodic and semantic affinity in the base embedding space. After applying the trained adapter, the magnitude of $\Delta$ increases and becomes store-consistent: episodic queries exhibit more negative $\Delta_{\text{after}}$, whereas semantic queries exhibit more positive $\Delta_{\text{after}}$. This suggests that the adapter reshapes the embedding space to increase the relative proximity between queries and the memory store they predominantly require.
\onecolumn
\raggedbottom
\section{Prompts Used}
\begin{tcolorbox}[colback=blue!5!white,colframe=blue!75!black, title=Prompt For Answer Generation]

You are a reasoning assistant tasked with answering questions based on retrieved conversational memory. \\

You are given: \\
• A question \\
• The top-5 retrieved texts from a long conversation \\
  • Each text may be: \\
    • A dialogue entry between two participants, OR \\
    • An observation derived from the dialogue, OR \\
    • A dialogue entry from user to a chat assistant \\

Your task is to: \\
1. Carefully read all retrieved texts. \\
2. Identify the information relevant to the question. \\
3. Combine evidence across multiple texts if needed. \\
4. Resolve references (e.g., pronouns, implicit mentions). \\
5. If conflicting information appears, prefer the most recent or most specific evidence.\\
6. Return a precise concise answer. If the question cannot be answered using the retrieved texts, respond with:
   "Insufficient information." \\

\textbf{Question}: \{question\}\\
\textbf{Retrieved Texts}: \{List of Retrieved Texts\} \\

\textbf{Output Format:} \\
\textbf{Answer}: <final answer> \\
\end{tcolorbox}
\begin{tcolorbox}[colback=blue!5!white,colframe=blue!75!black,title=Prompt For LLM-As-a-Judge]
You are an expert language model evaluator. I will provide you with a question, a
ground-truth answer, and a model-generated response. Your task is to determine whether
the response correctly answers the question by following these evaluation rules:\\
  • Answer Yes if the response contains or directly matches the correct answer.\\
  • Answer No if the response provides only a partial answer or omits essential information.\\
  • Answer No if the response does not sufficiently address the question.

Examples:

Example 1: Correct Response\\
• Question: What is the capital of France?\\
• Ground-truth Answer: Paris\\
• Response: The capital of France is Paris.\\
Evaluation:\\
• Output: Yes\\

Example 2: Incorrect Response \\
• Question: What is the capital of France? \\
• Ground-truth Answer: Paris \\
• Response: France is a country in Europe. \\
Evaluation:\\
• Output: No\\

\textbf{Additional Instructions:}\\
• Apply the evaluation criteria consistently. \\
• Base your decision strictly on the information in the response. \\
• Avoid subjective interpretations and adhere to the provided examples. \\
\textbf{Input:}\\
• \textbf{Question:} \{question\} \\
• \textbf{Ground-truth Answer:} \{gold\_answer\} \\
• \textbf{Response:} \{answer\} \\
\textbf{Output:}
\end{tcolorbox}
\begin{tcolorbox}[colback=blue!5!white,colframe=blue!75!black,title=Prompt For LLM Based Citations]
\textbf{Task Description}: Given a user query and a list of memories consisting of personal summaries with their corresponding original turns, generate a natural and fluent response while adhering to the following guidelines:\\
• Cite useful memories using [i], where i corresponds to the index of the cited memory.\\
• Do not cite memories that are not useful. If no useful memory exist, output [NO\_CITE].\\
• Each memory is independent and may repeat or contradict others. The response must be directly supported by cited memories.\\
• If the response relies on multiple memories, list all corresponding indices, e.g., [i, j, k].\\
• The citation is evaluated based on whether the response references the original turns, not the summaries.\\

\textbf{Examples}:

Case 1: Useful Memories Found\\
INPUT:\\
• User Query: What kind of work do I do and do I have any pets?\\
• Memories:\\
- Memory [0]:
 Speaker 1: Please make sure not to order anything with peanuts, I'm severely allergic.\\
- Memory [1]:
 Speaker 1: I need to head home a bit early today to take Max to the vet.\\
- Memory [2]:
 Speaker 1: I just landed a huge client for a new logo and branding project!\\

Output:\\
Answer: You work as a freelance graphic designer and you have a golden retriever named Max.\\
Citations: [1, 2]\\

Case 2: No Useful Memories\\
INPUT:\\
• User Query: What is my favorite movie?\\
• Memories:\\
- Memory [0]:
 Speaker 1: Please make sure not to order anything with peanuts, I'm severely allergic.\\
- Memory [1]:
 Speaker 1: I need to head home a bit early today to take Max to the vet.\\
- Memory [2]:
 Speaker 1: I just landed a huge client for a new logo and branding project!\\
 Speaker 1: I just landed a huge client for a new logo and branding project!\\

Output:\\
Answer: I don't have enough information to answer that.\\
Citations: [NO\_CITE]\\

\textbf{Additional Instructions}:
• Ensure the response is fluent and directly answers the user's query.\\
• Always cite the useful memory indices explicitly.\\
• The citation is evaluated based on whether the response references the original turns, not the summaries.\\
• Follow the format of the examples provided above.\\

\textbf{Input}:\\
• User Query: \{query\}\\
• Memories: \{memories\}\\
\textbf{Output}:\\
\end{tcolorbox}
\end{document}